\documentclass[11pt, twocolumn]{article}
\usepackage[a4paper, margin=1.5cm]{geometry}
\usepackage{chemformula}
\usepackage[T1]{fontenc}
\usepackage{amsmath}
\usepackage{amssymb}
\usepackage{amsfonts}
\usepackage{graphicx}
\usepackage[labelfont=bf]{caption}
\usepackage{bbm}
\usepackage{booktabs}
\usepackage{multirow}
\usepackage{longtable}
\usepackage{makecell}
\usepackage{adjustbox}
\usepackage{authblk}
\usepackage{placeins}
\usepackage[superscript]{cite}
\usepackage{url}
\usepackage[many]{tcolorbox}
\usepackage{soul}
\usepackage{wasysym}
\usepackage{CJKutf8}
\usepackage{hyperref}

\usepackage{floatrow}
\usepackage[label font=bf,labelformat=simple]{subfig}
\usepackage{caption}
\newcommand{\syntheseus}{syntheseus}

\newcommand{\smilesmodel}{R-SMILES 2}
\newcommand{\locmodel}{NeuralLoc}

\newcommand{\combinedmodel}{RetroChimera}

\title{Chemist-aligned retrosynthesis\\by ensembling diverse inductive bias models}

\author{Krzysztof Maziarz*$^1$,
Guoqing Liu (\begin{CJK*}{UTF8}{gbsn}刘国庆\end{CJK*})*$^1$,
Hubert Misztela$^2$,
Austin Tripp$^3$,
Junren Li$^1$,
Aleksei Kornev$^2$,
Piotr Gaiński$^{1,4}$,
Holger Hoefling$^2$,
Mike Fortunato$^2$,
Rishi Gupta$^2$,
Marwin Segler$^1$
\\
\quad
\\
$^1$Microsoft Research AI for Science; $^2$Novartis Biomedical Research;\\
$^3$University of Cambridge;
$^4$Jagiellonian University;
*Equal core contributor\\
Correspondence to \texttt{\{krmaziar,guoqingliu,marwinsegler\}@microsoft.com}
\vspace{-0.25cm}
}

\date{}

\begin{document}

\maketitle

\begin{abstract}
Chemical synthesis remains a critical bottleneck in the discovery and manufacture of functional small molecules.
AI-based synthesis planning models could be a potential remedy to find effective syntheses, and have made progress in recent years. However, they still struggle with less frequent, yet critical reactions for synthetic strategy, as well as hallucinated, incorrect predictions. This hampers multi-step search algorithms that rely on models, and leads to misalignment with chemists' expectations. 
Here we propose \combinedmodel: a frontier retrosynthesis model, built upon two newly developed components with complementary inductive biases, which we fuse together using a new framework for integrating predictions from multiple sources via a learning-based ensembling strategy.
Through experiments across several orders of magnitude in data scale and splitting strategy, we show \combinedmodel~outperforms all major models by a large margin, demonstrating robustness outside the training data, as well as for the first time the ability to learn from even a very small number of examples per reaction class.
Moreover, industrial organic chemists prefer predictions from \combinedmodel~over the reactions it was trained on in terms of quality, revealing high levels of alignment.
Finally, we demonstrate zero-shot transfer to an internal dataset from a major pharmaceutical company, showing robust generalization under distribution shift. 
With the new dimension that our ensembling framework unlocks, we anticipate further acceleration in the development of even more accurate models.
\end{abstract}

\section*{Introduction}
Chemical Synthesis is central to the discovery and supply of small molecule-based therapeutics, materials, and fine chemicals. However, as syntheses often fail, and thus constitute a critical bottleneck, using computational methods to propose better synthesis routes is highly desirable.\cite{stanley2023fake, shields2024aizynth, tu2023predictive}
Computer-aided synthesis planning has a long research history, with tools traditionally implemented via rule-based expert systems.\cite{vleduts1963concerning,corey1969computer,ihlenfeldt1996computer} However, over several decades progress had been limited.\cite{ihlenfeldt1996computer} Since 2017, significant advancements have been made, along two directions. First, the expert system approach of manually coding reaction rules has been reimplemented~\cite{hastedt2024investigating, schwaller2022machine} by Szymkuc and coworkers, and has been experimentally validated.\cite{klucznik2018efficient,mikulak2020computational} 
Second, by re-framing synthesis planning as a machine learning (ML) problem, where deep neural networks are trained on large reaction datasets to predict synthetic disconnections and reaction outcomes, which are then coupled with neural-guided search, a paradigm shift has been achieved~\cite{segler2017neural, segler2018planning, coley2019robotic}.
Since then, several new ML models~\cite{liu2017retrosynthetic, schwaller2019molecular, dai2019retrosynthesis, tetko2020state, sacha2021molecule, chen2021deep, tu2022permutation, zhong2022root, igashov2023retrobridge,wang2023retrosynthesis,  xie2023retrosynthesis, laabid2024alignment, gainski2024retrogfn, westerlund2024chemformers, zhang2024retrosynthesis, han2024retrosynthesis, deng2025rsgpt} and search algorithms~\cite{chen2020retro, xie2022retrograph, liu2023retrosynthetic, tripp2023retro} have been introduced. Incorporated into readily available tools for retrosynthetic search, which are increasingly used in computational workflows and as a source of inspiration for chemists during route planning, ML-based synthesis planning has also been experimentally validated.\cite{coley2019robotic, genheden2020aizynthfinder, shields2024aizynth, tu2025askcos}

\begin{figure*}[!ht]
    \centering
    \vspace{-0.25cm}
    \includegraphics[width=0.8\linewidth]{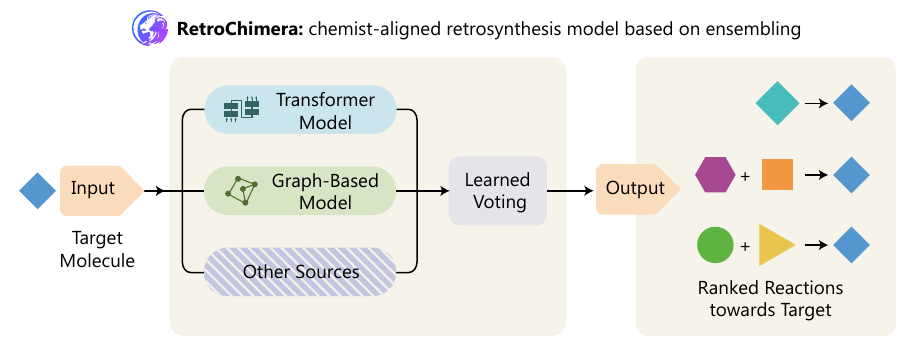}
    \vspace{-0.15cm}
    \caption{Our framework for ensemble-based retrosynthesis with learned reranking which underpins \combinedmodel. The ensemble receives a target molecule as the input, which is then processed by the constituent models. The model outputs are then aggregated using a learning-to-rank strategy. While in this work we only investigate deep learning models as prediction sources (solid boxes), it is possible to add additional sources, for example calls to reaction databases or human-in-the-loop queries, which will be addressed in future work (dashed box).}
    \label{fig:overview}
\end{figure*}

While conceptually ML-based synthesis planning promises favorable scaling with the ever-growing body of organic chemistry knowledge in the literature, patents, and electronic laboratory notebooks, so far, compared to hand-coded expert systems, ML-based planning suffered from limited accuracy in particular for rare reaction classes, limited robustness further away from the training distribution, and reduced acceptance by chemists.\cite{tu2023predictive} 
In addition, chemists often combine multiple strategies, from direct pattern matching to envisioning new transformations, which computational approaches currently do not reflect.

In this work, we present a framework for retrosynthesis prediction that ensembles models with diverse inductive biases using a learning-to-rank strategy.
Instantiated with two new state-of-the-art models, also introduced here -- one based on Graph Neural Networks using molecular edit rules and one on de-novo generation using a modern Transformer -- we obtain \combinedmodel, which achieves high accuracy on common and rare reactions alike, increased robustness, as well as superior performance in multi-step search. Furthermore, we show quantitatively that organic chemists prefer \combinedmodel~over reported reactions from the literature, and elucidate the ability of our probabilistic model to learn robustly even when presented with partially noisy training data.

\section*{Computer-Aided Synthesis Planning}
Systems for Computer-Aided Synthesis Planning usually perform retrosynthesis, i.e. predicting transformations which correspond to reverse chemical reactions starting with the target molecule, and have four components: (1) a single-step model or algorithm to propose transformations that correspond to feasible reactions in the forward direction, (2) a search algorithm that chains together transformations into multi-step routes, (3) ranking criteria for the routes, and (4) admissible building block molecules into which the target has to be deconstructed.\cite{strieth2020machine,tu2023predictive}
Thus, an accurate single-step model is crucial as it defines the search space of possible reactions to explore. As the model is called recursively during search, the requirements for accuracy are very strict, as errors compound with multiple steps, and a single error will invalidate the entire route. In addition, it is critical for the model to cover a large chemical reaction space, so that strategic yet rare transformations are not missed. 

Current single-step models can be classified into \textit{editing} models, which change only the parts of the molecule involved in the reaction, e.g. make or break bonds and add leaving groups, or \textit{de-novo} models, which generate the reactant structures from scratch, including regeneration of the unchanged parts.  
While in recent years several models have been proposed, high accuracy still poses a significant challenge, especially for reaction types of lower precedence.\cite{segler2018planning, fortunato2020data, seidl2022improving, hassen2022mind, mikulak2020computational,tanovic2025exploration} However, rarer reactions are often highly specific and strategically useful.\cite{mikulak2020computational} 

\section*{Ensembling}

Model ensembling is an ML technique where several models trained to perform the same task are combined to obtain better performance than any of them would in isolation~\cite{sewell2008ensemble}. Generally, ensembles work best when the models being combined are diverse.\cite{theisen2024ensembles} In retrosynthesis prediction, several options of ensembling exist. Instead of directly ensembling in token probability space, which can only be applied to autoregressive models, we can perform count-based ensembling in  molecule space by aggregating outputs shared by ensembled models, which we hypothesize to be more expressive.
Moreover, count-based ensembling is much more versatile, as it can ensemble any set of models, as well as non-model sources of reactions; for example, it would allow to mix in proposals coming from lookups in reaction databases, or manual input from chemists.

Here, we propose a strategy to merge several output lists based on overlaps between them, which for the first time leads to substantial gains over the ensembled models.

Given ranked outputs $r_{i,k}$ from $m$ models, where $r_{i,k}$ is the $k$-th top prediction from the $i$-th model, we rank unique reactant sets $r$ by decreasing $\texttt{score}(r)$:

\begin{equation}\label{eq:ensemble}
\texttt{score}(r) = \sum_{i = 1}^m \sum_{k = 1}^{k_{max}} \mathbbm{1}{[r = r_{i, k}]} \cdot \theta_{i, k},
\end{equation}

where $k_{max}$ is maximum number of predictions considered per model and $\theta \in \mathbb{R}_{+}^{m \times k_{max}}$; we omit the dependence of $\texttt{score}$ and losses defined below on $\theta$ for clarity.
In other words, reactant set predicted at rank $k$ by model $i$ is assigned score $\theta_{i, k}$, with scores summed across models. Intuitively, reactions ranking highly across several models will be assigned a larger score than those suggested by a single model.

Inspired by work on learning to rank~\cite{burges2005learning}, we learn $\theta$ from predictions on the validation set $\mathcal{D}_{val}$ by minimizing

\begin{equation}\label{eq:lrank}
\mathcal{L}_{rank} = \mathbb{E}_{(p, r^+) \in \mathcal{D}_{val}} \sum_{r^- \in \mathcal{R}^-} \mathcal{L}_{rank}(r^+, r^-),
\end{equation}

\begin{equation*}
\mathcal{L}_{rank}(r^+, r^-) = \sigma \left(\frac{\texttt{score}(r^-) - \texttt{score}(r^+) + \epsilon}{T} \right),
\end{equation*}

where $\mathcal{R}^- = \{r_{i,k}: r_{i,k} \neq r^+\}$ are predictions differing from ground-truth $r^+$ and $\epsilon$ is a small constant. For $\epsilon, \ T \rightarrow 0$, $\mathcal{L}_{rank}(r^+, r^-) \rightarrow \mathbbm{1}{[\texttt{score}(r^-) > \texttt{score}(r^+)]}$, i.e. indicator of whether $r^+$ and $r^-$ are ordered incorrectly. In the limit $\mathcal{L}_{rank}$ lacks useful gradients, thus we start with $T > 0$ and linearly anneal to $0$ over the course of optimization. To avoid overfitting to $\mathcal{D}_{val}$ we constrain each $\theta_i$ to be decreasing and convex.

In the experiments we optimize $\theta$ on the validation set and evaluate on the test set; see Methods for more details and hyperparameters of this procedure.
We find that our strategy consistently outperforms other approaches, and learns non-trivial schemes where relative model importance depends on $k$ (Extended Data Figure~\ref{fig:ensemble_analysis}).

\paragraph{Ensembling public models}

To test our strategy, we consider models trained on USPTO-50K available in \syntheseus~\cite{maziarz2023re}: Chemformer~\cite{irwin2022chemformer}, GLN~\cite{dai2019retrosynthesis}, Graph2Edits~\cite{zhong2023retrosynthesis}, LocalRetro~\cite{chen2021deep}, MEGAN~\cite{sacha2021molecule}, RetroKNN~\cite{xie2023retrosynthesis} and R-SMILES~\cite{zhong2022root}. We also retrain R-SMILES to study the effect of ensembling two instances of the same model, and include our reimplementation of Edit Rule Classification (``NeuralSym'')~\cite{segler2017neural}.
Remarkably, ensembling any pair of models results in performance better than attained by either (Extended Data Figure~\ref{fig:ensemble}), even when combining a strong model with a weaker one: for example, top-5 accuracy of R-SMILES can be improved by 1.5\% by ensembling with GLN, despite it being significantly weaker.
However, models employing similar modeling show limited benefit from being combined, which suggests diversity is key to a strong ensemble, and motivates us to propose \emph{two} models -- one based on molecule editing and one on de-novo generation -- and investigate the performance of their ensemble at scale. Prior work often deems ensembles incomparable to individual models due to higher cost~\cite{schwaller2019molecular}, but we challenge this assumption noting that ensembling a fast editing model with a de-novo Transformer leads to a negligible cost increase over the latter. In the following sections, we introduce our models, and benchmark them at increasing data scales.

Ensembles discussed above already set a new state of the art on USPTO-50K, even outperforming model-reranker combinations~\cite{lin2022improving}. However, in the following sections we show even better performance by utilizing our newly proposed models.

\section*{Model architecture}

\begin{figure*}[!ht]
    \centering
    \sidesubfloat[]
    {\includegraphics[width=0.95\linewidth]{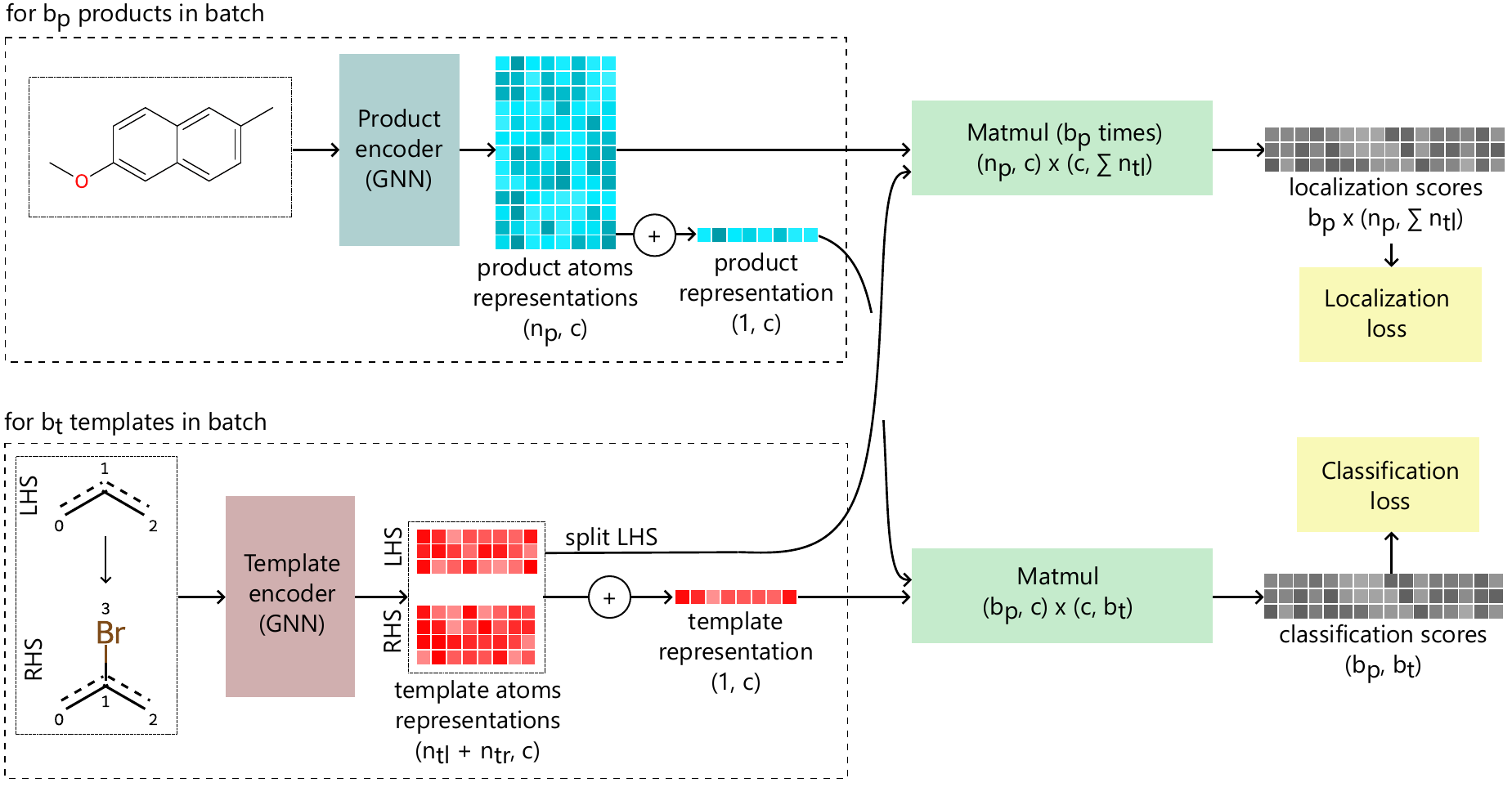}}
    \vspace{0.2cm}\\
    \hspace{-1.5cm}
    \sidesubfloat[]{\hspace{0.3cm}\includegraphics[scale=0.5]{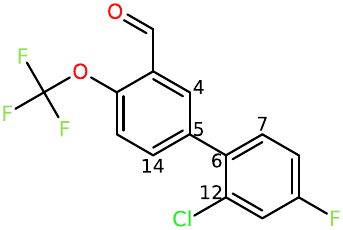}}
    \hfil
    \sidesubfloat[]{\hspace{0.3cm}\includegraphics[scale=0.5]{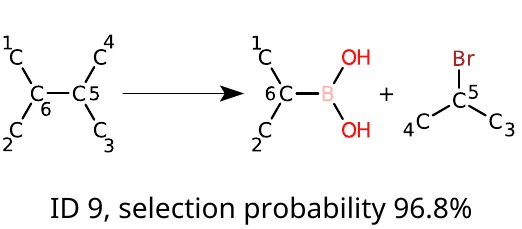}}
    \hfil
    \sidesubfloat[]
    {\includegraphics[height=4.7cm]{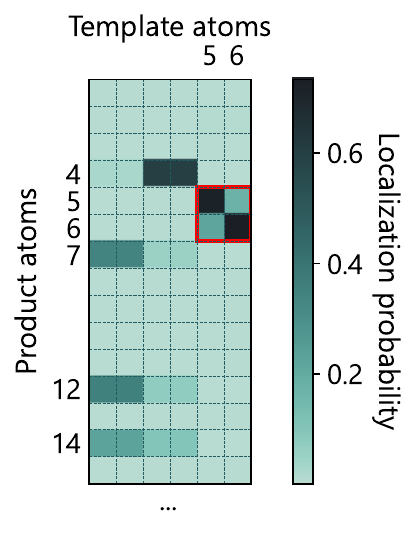}}\\
    \vspace{-2.2cm}\hspace{-7.6cm}
    \sidesubfloat[]{\hspace{0.3cm}\includegraphics[scale=0.5]{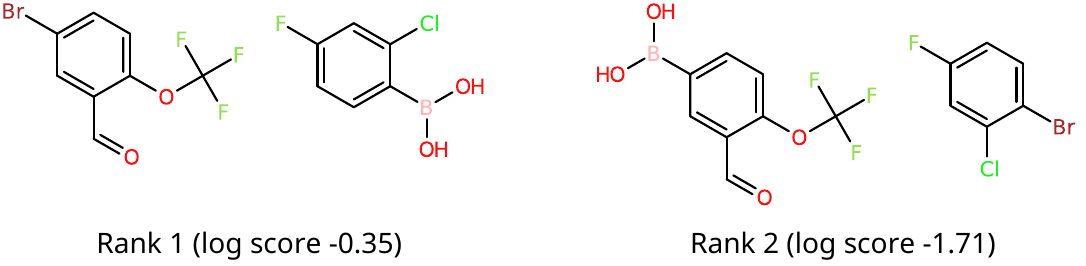}}\\
    \caption{\textbf{a}, Architecture of the editing model (\locmodel). Product and templates are encoded through Graph Neural Network encoders to produce contextualized atom representations. 
    Template scores are computed by multiplying product representation with template representations. Localization scores are computed as products of product atom representations and template left-hand side atom representations. All templates in the batch are used for classification, but only a subset is used for localization for a given product.
    \textbf{b-e}, Inference process. \textbf{b}, Product is input into the network (atom IDs are not part of model input; shown to contextualize the localization). \textbf{c}, Classification head selects a template from the library. \textbf{d}, Product and template atom representations determine localization scores (shown for first 15 atoms). \textbf{e}, As the template is symmetric, application produces two reactant sets depending on how the \textsc{C:5-C:6} bond is matched. Localization differentiates them, suggesting to match \textsc{C:5} in the product with \textsc{C:5} in the template (red square in \textbf{d}). This proceeds for several top templates; resulting reactants are ranked based on a combination of classification and localization. In this case, \locmodel~correctly prefers the result that is more chemically plausible.}
    \label{fig:loc_model}
\end{figure*}

We instantiate \combinedmodel~as an ensemble of two separately trained models -- one based on molecule editing and one on de-novo generation -- each designed to address specific limitations in their respective modeling classes.
As the edit-based model can be implemented very efficiently, \combinedmodel~delivers inference cost comparable to a single de-novo model such as R-SMILES, however -- as seen in the later sections -- with superior predictive performance.

\paragraph{Editing Model}

Molecule-editing models tend to stay closer to the data distribution due to reliance on symbolic transformations that have support in training data, especially when the edits are limited to stricter reaction rules or templates.
Even though they were the first ML-based retrosynthesis model, template classification continues to be a default choice in contemporary workflows.
However, two limitations hinder these models at scale: (1) weights responsible for choosing the template are treated as free parameters, precluding representational transfer between templates; and (2) applying a template can produce more than one prediction due to multiple matches in the input molecule, and these alternatives are not differentiated. Prior work has explored partial solutions: (1) by using a template encoder~\cite{seidl2022improving}; and (2) by separately predicting the reaction centre to constrain template match~\cite{dai2019retrosynthesis,chen2021deep,sacha2023molecule} or by introducing a separate module to rank the final reactant sets~\cite{dai2019retrosynthesis}. However, narrowing template application to the reaction centre may not be enough to uniquely specify the reactants due to symmetry (Figure~\ref{fig:loc_model}c).

Inspired by these works we design \locmodel, a new template classification model (Figure~\ref{fig:loc_model}a). Apart from a product encoder, \locmodel~contains a separate template encoder; unlike MHNreact~\cite{seidl2022improving}, this encoder directly processes the template as a graph using a tailored featurization (Methods).
Our model uses aggregated product and template representations for template classification, and atom-level representations for localization by computing pairwise assignment probabilities between product and template atoms.
During inference (Figure~\ref{fig:loc_model}b-e) we call the classification branch, apply a number of top-scoring templates, and reorder all results taking localization into account; see Methods for architectural details, hyperparameters, and description of model training and inference.

\paragraph{De-Novo Model}

\begin{figure*}[!ht]
    \centering
    \includegraphics[width=0.85\linewidth]{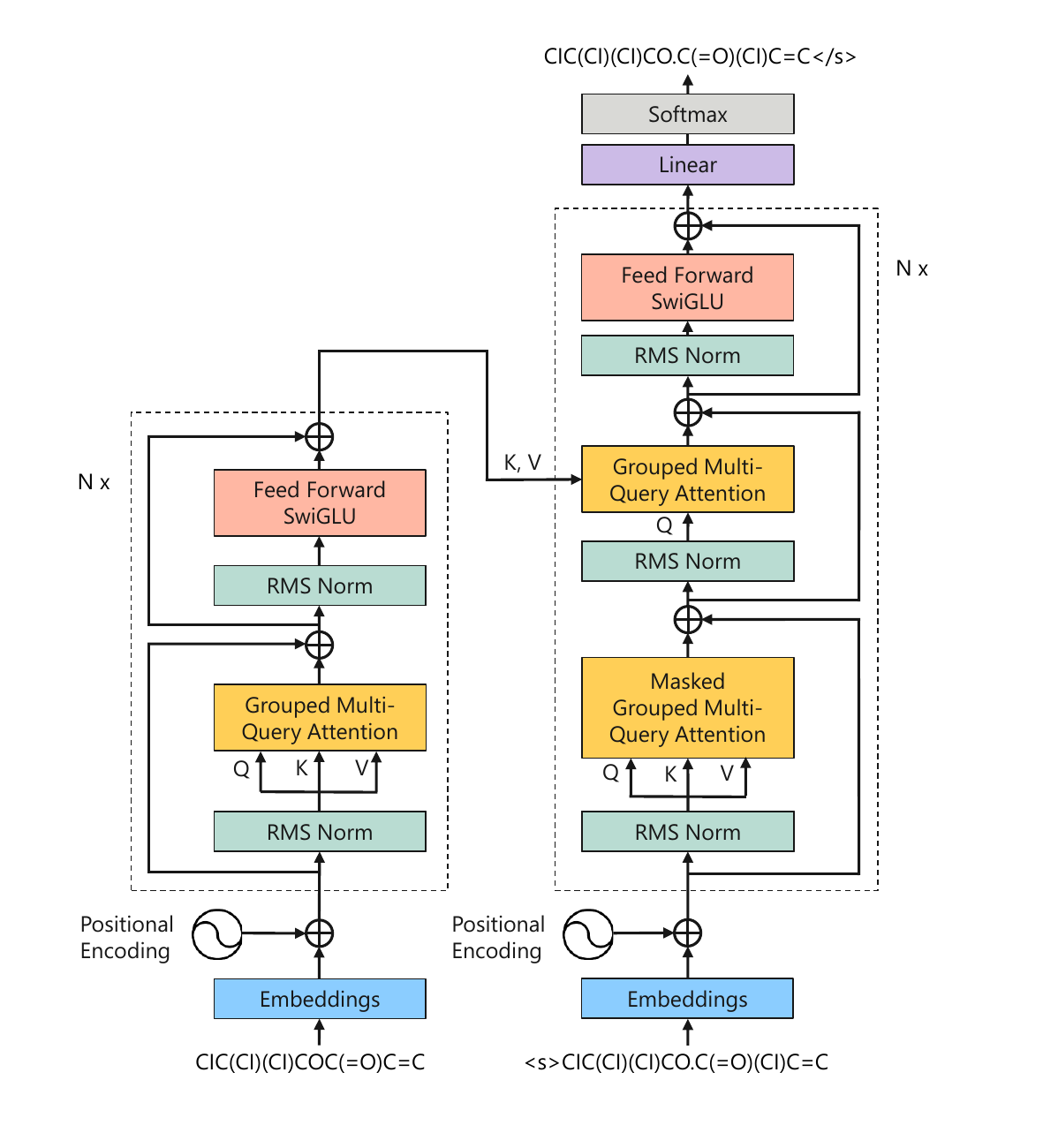}
    \caption{Architecture of the de-novo model (\smilesmodel). The input product is converted to a SMILES string and tokenized into a sequence of tokens. Before the sequence is processed further, sinusoidal positional embeddings are incorporated to infuse positional information. The sequence then undergoes transformation through layers composed of grouped multi-query attention, RMS normalization, and feedforward layers with SwiGLU activations. The autoregressive decoder predicts the SMILES sequence of reactants utilizing self-attention over already produced tokens and cross-attention over encoder output. The model is trained using a cross-entropy loss.}
    \label{fig:transformer_model}
\end{figure*}

We build our new de-novo model upon the Seq2Seq framework pioneered by Liu et al~\cite{liu2017retrosynthetic}, and the successful R-SMILES model~\cite{schwaller2019molecular, zhong2022root}, 
which utilizes an aligned SMILES format to represent input products and ground-truth reactants.
This involves training an encoder-decoder model based on the Transformer architecture~\cite{vaswani2017attention, touvron2023llama2openfoundation, jiang2023mistral7b} using a cross-entropy loss.
Unlike previous work relying on OpenNMT~\cite{klein-etal-2017-opennmt}, 
we employ three architectural modifications to improve accuracy and inference speed:
(1) Group-Query Attention (GQA)~\cite{ainslie2023gqa} instead of standard multi-head attention to reduce computational complexity; (2) pre-normalization using RMSNorm~\cite{zhang2019rms} instead of LayerNorm; and (3) SwiGLU activation~\cite{shazeer2020glu} instead of ReLU in feedforward layers.
We also refined the beam search termination condition to better suit the domain, improving top-$k$ accuracy for large $k$.
We refer to our updated model as \smilesmodel~(Extended Data Figure~\ref{fig:transformer_model}); see Methods for more details.

\section*{Results on reaction prediction}
To investigate the performance of our framework and models, we start with small-scale experiments on USPTO-50K, and then scale to the largest available public dataset and better curated in-house datasets.

\paragraph{USPTO}

\looseness=-1
For a comparison on public data we use USPTO-50K and USPTO-FULL datasets preprocessed by prior work~\cite{dai2019retrosynthesis}.
We follow best evaluation practices~\cite{maziarz2023re} and use \syntheseus~to benchmark our models as well as those baselines that are integrated into the library. We selected the baselines to include best performing methods while avoiding juxtaposing results obtained on different dataset versions; see Methods for further discussion.

We find that \locmodel~and \smilesmodel~generally match or surpass the state of the art within their own model classes, while \combinedmodel~performs better than both and sets new state of the art for $k > 1$ on both USPTO-50K and USPTO-FULL, pushing the top-10 accuracy by $1.7\%$ and $1.6\%$, respectively (Extended Data Figure~\ref{fig:accuracy_uspto}, Extended Data Tables~\ref{tab:uspto-50k} and \ref{tab:uspto-full}).
To test the scaling of our ensembling strategy, we also evaluated an ensemble containing both our proposed models and most of the baselines, and found it pushes the state of the art even further, although it may not be practical due to excessive resource requirements.
Nevertheless, this result may inspire future work on model distillation.

To obtain a good trade-off between resource requirements and accuracy, we focus on ensembling two models and scale \combinedmodel~to larger and more diverse datasets.

\paragraph{Pistachio}

\newcommand{\plotheight}{7.35cm}

\begin{figure*}[ht!]
    \hspace{-0.2cm}
    \sidesubfloat[]{\hspace{-0.45cm}\includegraphics[height=\plotheight]{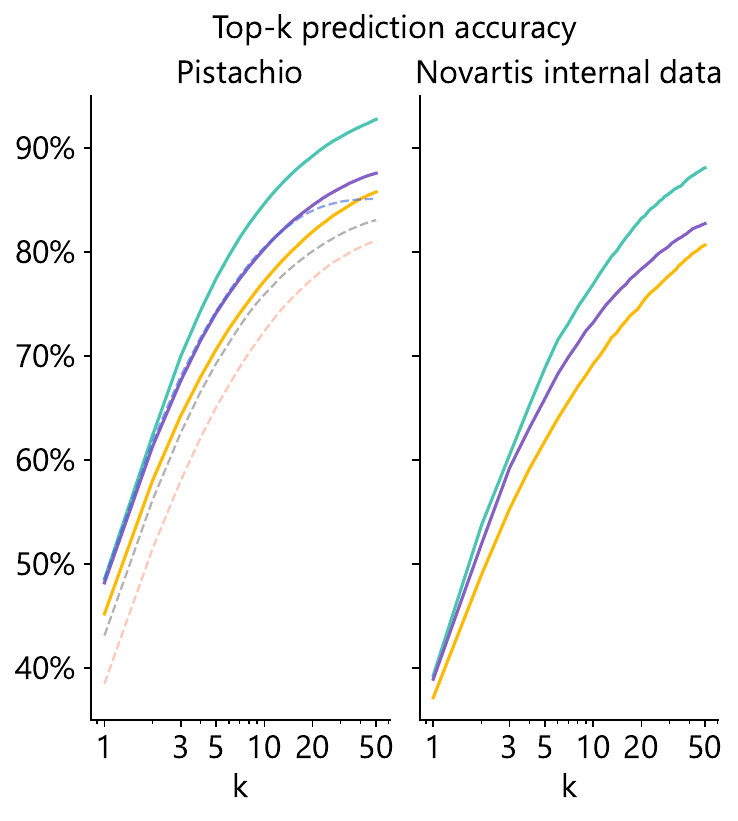}\label{fig:accuracy_pistachio_nvs}}
    \hfil
    \hspace{-0.2cm}
    \sidesubfloat[]{\hspace{-0.45cm}\includegraphics[height=\plotheight]{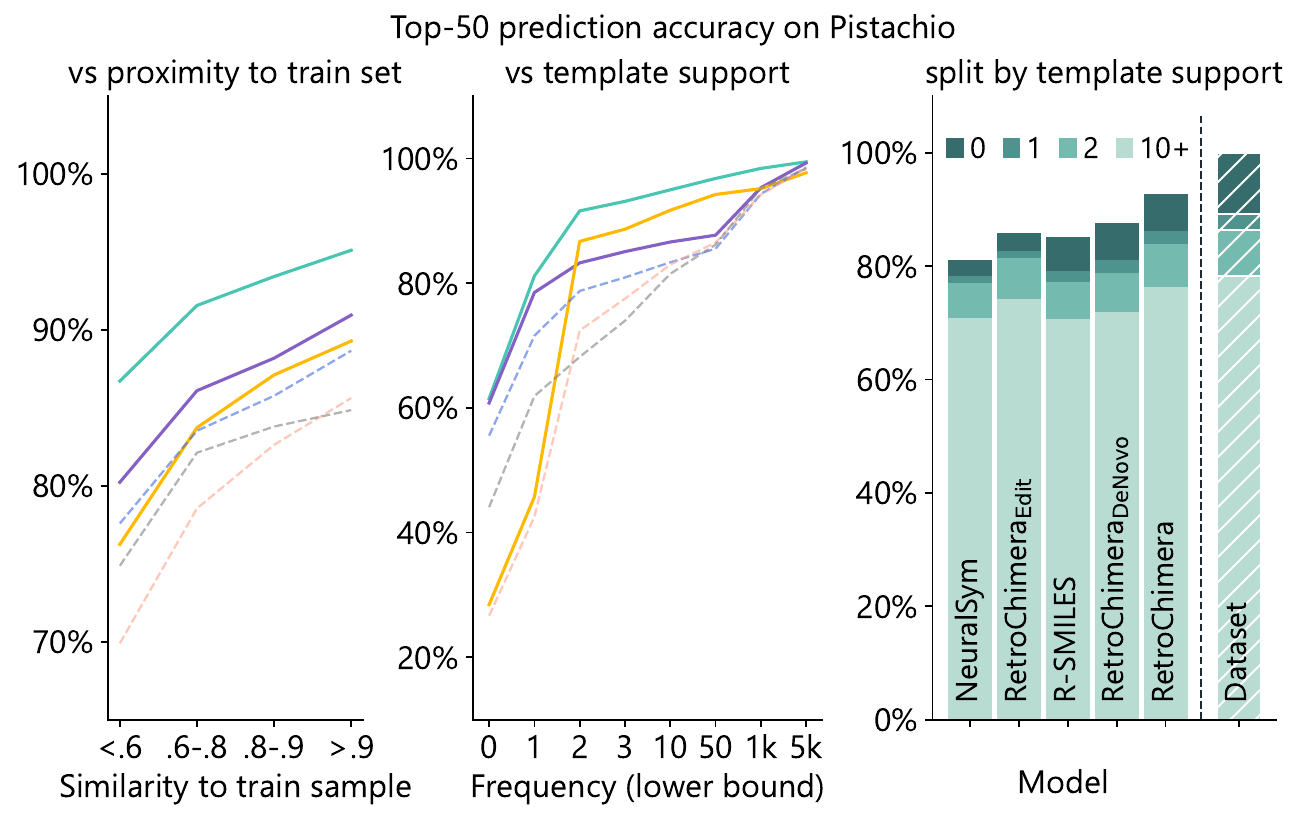}\label{fig:accuracy_vs_similarity}}\\
    \includegraphics[height=0.55cm]{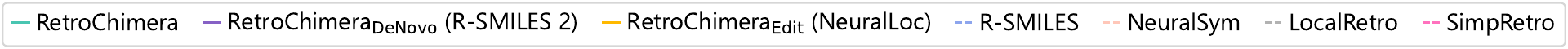}\\

    \hspace{-0.2cm}
    \sidesubfloat[]{\hspace{-0.45cm}\includegraphics[height=\plotheight]{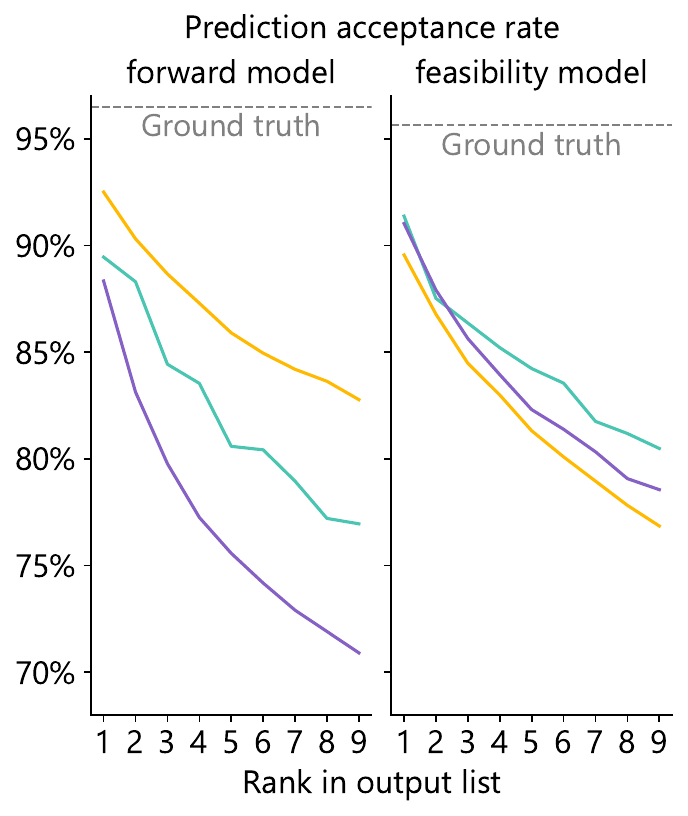}\label{fig:forward_and_feasibility}}
    \hfil
    \sidesubfloat[]{\hspace{-0.28cm}\includegraphics[height=\plotheight]{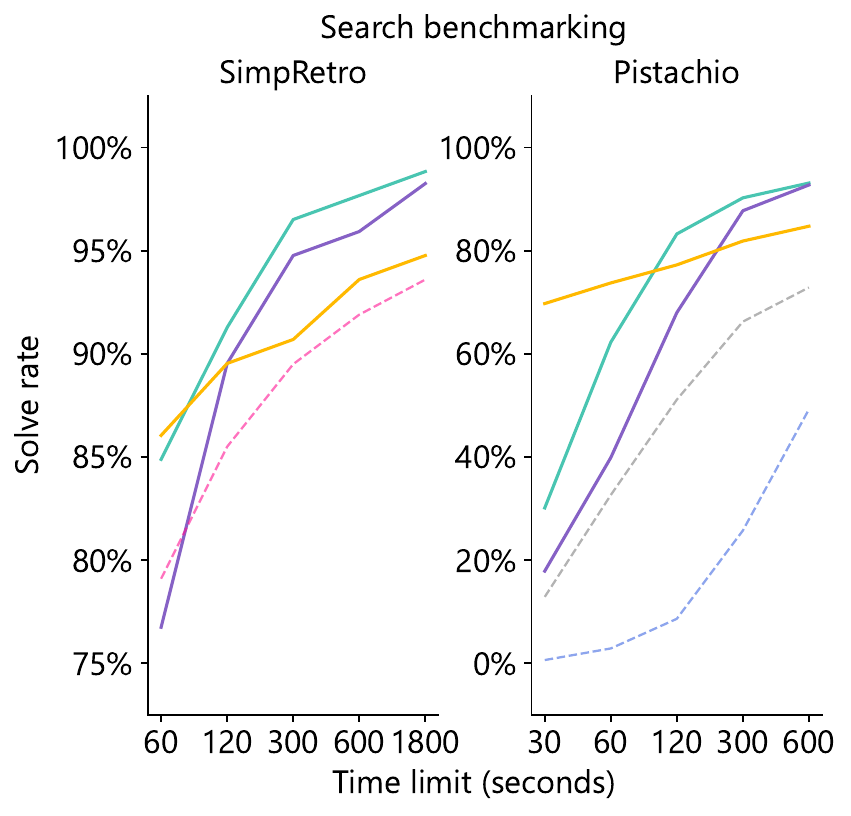}\label{fig:search_solve_rate}}
    \hfil
    \sidesubfloat[]
    {\hspace{-0.3cm}\includegraphics[height=\plotheight]{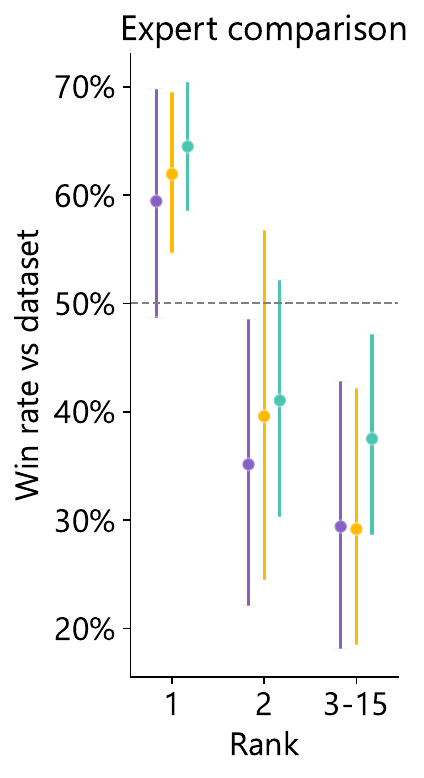}
    \label{fig:preference}}
    \caption{Benchmarking Pistachio-trained models (ours shown as solid lines, baselines as dashed).
    \textbf{a}, Accuracy on Pistachio (left) and Novartis data (right). 
    \textbf{b}, Top-50 accuracy on Pistachio when grouping by Morgan fingerprint similarity (Tanimoto, radius 2) to a training product (left) or template frequency (middle, right).
    \textbf{c}, Fraction of non-ground-truth predictions accepted by forward (left) and feasibility (right) models, as a function of rank; dashed line shows the acceptance rate of dataset ground-truths.
    \textbf{d}, Solve rate on the SimpRetro dataset (left) and on hard products from Pistachio (right).
    \textbf{e}, Win rate against dataset ground-truth conditioned on the prediction being different from the dataset, estimated from pairwise expert comparison data. Whiskers correspond to 95\% confidence interval from $1000$ bootstrap resamples.}
\end{figure*}

We scale our models to the proprietary Pistachio dataset, which is better curated and represents a 3.5x increase in number of samples compared to USPTO-FULL.
We use the data prepared by Maziarz et al~\cite{maziarz2023re}, where reactions present in the database as of June 2023 were grouped by product and randomly split into three folds.
We reuse the training and validation sets, and build a new time-split test set: we take reactions added to Pistachio in 2024, marked as high quality by the database curator, and whose product had fingerprint similarity to a training product below $0.95$ (see Methods).
This gave rise to a high quality test set of $146\,393$ reactions both temporally and structurally separate from the data used for training and validation; we use it as our default test set and defer the results on the original test set to Extended Data Figure~\ref{fig:accuracy_pistachio_random}.
As there are no published results on this version of Pistachio, we also train and evaluate selected, strong baselines (LocalRetro, R-SMILES, NeuralSym).

Similarly to the results on USPTO, our models establish state-of-the-art performance within their respective classes (Figure~\ref{fig:accuracy_pistachio_nvs}).
\combinedmodel~matches \smilesmodel~for small $k$ while outperforming it for larger $k$ due to the pooling of diverse inductive biases.
With only $10$ results, \combinedmodel~reaches the accuracy of considering $50$ results from R-SMILES.

To further understand the strengths of the individual models, we analysed top-50 recall as a function of fingerprint similarity to training data, as well as frequency of the ground-truth template (Figure~\ref{fig:accuracy_vs_similarity}, see Methods for details).
All models perform better on reactions more similar to the training data, or those utilizing more common templates.
Far from training data de-novo models degrade less than edit-based ones, giving credence to a hypothesis that the former generalize better~\cite{tu2022retrosynthesis, westerlund2024chemformers}.
While \smilesmodel~outperforms \locmodel~on reactions with little to no template precedence, for moderate template support the trend reverses, showing that our editing model can use a template effectively from just a few examples.

When the models are combined into \combinedmodel, their complementary inductive biases lead to superior performance for both frequent and rare reaction types alike, effectively addressing the "rare reactions problem".
Moreover, \combinedmodel~reaches close to optimal recall on well-precedented reactions, indicating the model can be seen as a ``soft reaction database''. 

\paragraph{Transfer to internal data}

To assess performance on real world medicinal chemistry, we test our models on a dataset of $10\,444$ reactions extracted from an internal database maintained by Novartis. This is conducted in a zero-shot setting, where no additional finetuning of the models is performed, and ensembling coefficients $\theta$ are kept as determined on Pistachio.
While top-1 accuracy is comparatively lower compared to Pistachio (Figure~\ref{fig:accuracy_pistachio_nvs}), which could be due to distribution shift, \combinedmodel~still delivers the best performance, outperforming both \smilesmodel~and \locmodel.
This highlights that \combinedmodel~can be used beyond its original data distribution with no extra training, and bring strong performance in a real-world setting. We anticipate performance would likely improve with further fine-tuning.

\begin{figure*}[!h]
    \centering
    \vspace{0.25cm}
    \includegraphics[width=0.95\linewidth]{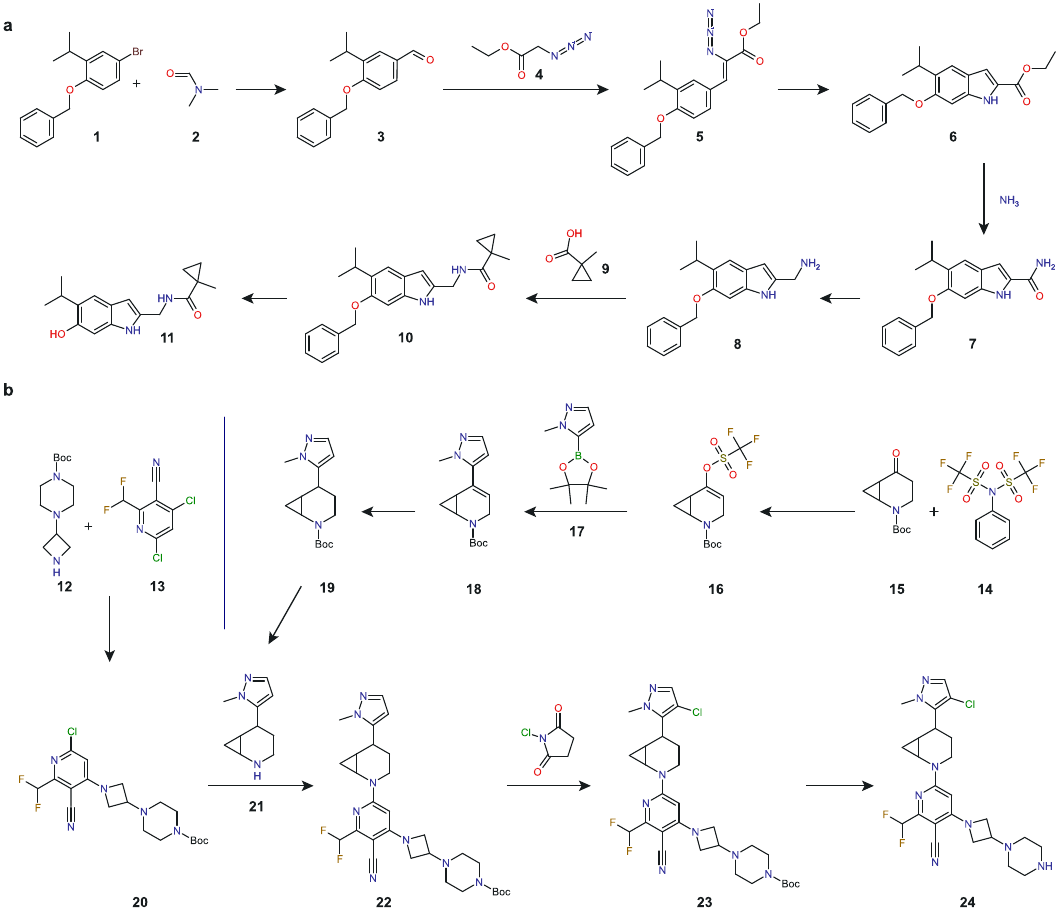}
    \caption{Example routes identified by \combinedmodel. Targets were selected from the Pistachio test set, and represent commonly observed challenges in medicinal chemistry synthesis. Note that in route \textbf{a} (from \textbf{5} to \textbf{6}) the model proposes to use a less frequent Hemetsberger–Knittel indole synthesis, which highlights the ability of the model to also propose reasonable reactions that chemists would likely not immediately think of. As reagents, solvents and reaction conditions were not predicted in this study, they were omitted from the depiction. \texttt{Boc} is tert-butyloxycarbonyl.}
    \label{fig:exampleroutes}
\end{figure*}

\paragraph{Reaction quality}

Accuracy tests how well a model can recall the ground-truth, but not whether its non-ground-truth predictions are reasonable, which is arguably more important for search~\cite{maziarz2023re}.
To assess how feasible model outputs are overall, it is common to either feed the predicted reactants to a forward model to measure round-trip accuracy~\cite{schwaller2019evaluation, chen2021deep}, or feed entire reactions to a feasibility model~\cite{gainski2024retrogfn}.
In general, feasibility models are preferred as those are trained with both positive and negative reactions, and can handle cases where the reactants would not react~\cite{maziarz2023re}.
Here we explore both routes: we use a forward model based on the \smilesmodel~architecture and a feasibility model based on the approach of Gaiński et al~\cite{gainski2024retrogfn}.
Both were trained on Pistachio and calibrated to accept around $95\%$ of ground-truths; see Methods for details.

We compute acceptance rate for each model and rank (Figure~\ref{fig:forward_and_feasibility}).
Interestingly, the scoring models partially disagree: both consider \combinedmodel~of higher quality than \smilesmodel, but the forward model judges \locmodel~much more highly.
This highlights that while the two scoring approaches correctly distinguish generated predictions from ground-truths, they leverage disparate heuristics.

\section*{Results on multi-step search}

\begin{figure*}[ht!]
    \vspace{0.4cm}
    \includegraphics[width=0.995\linewidth]{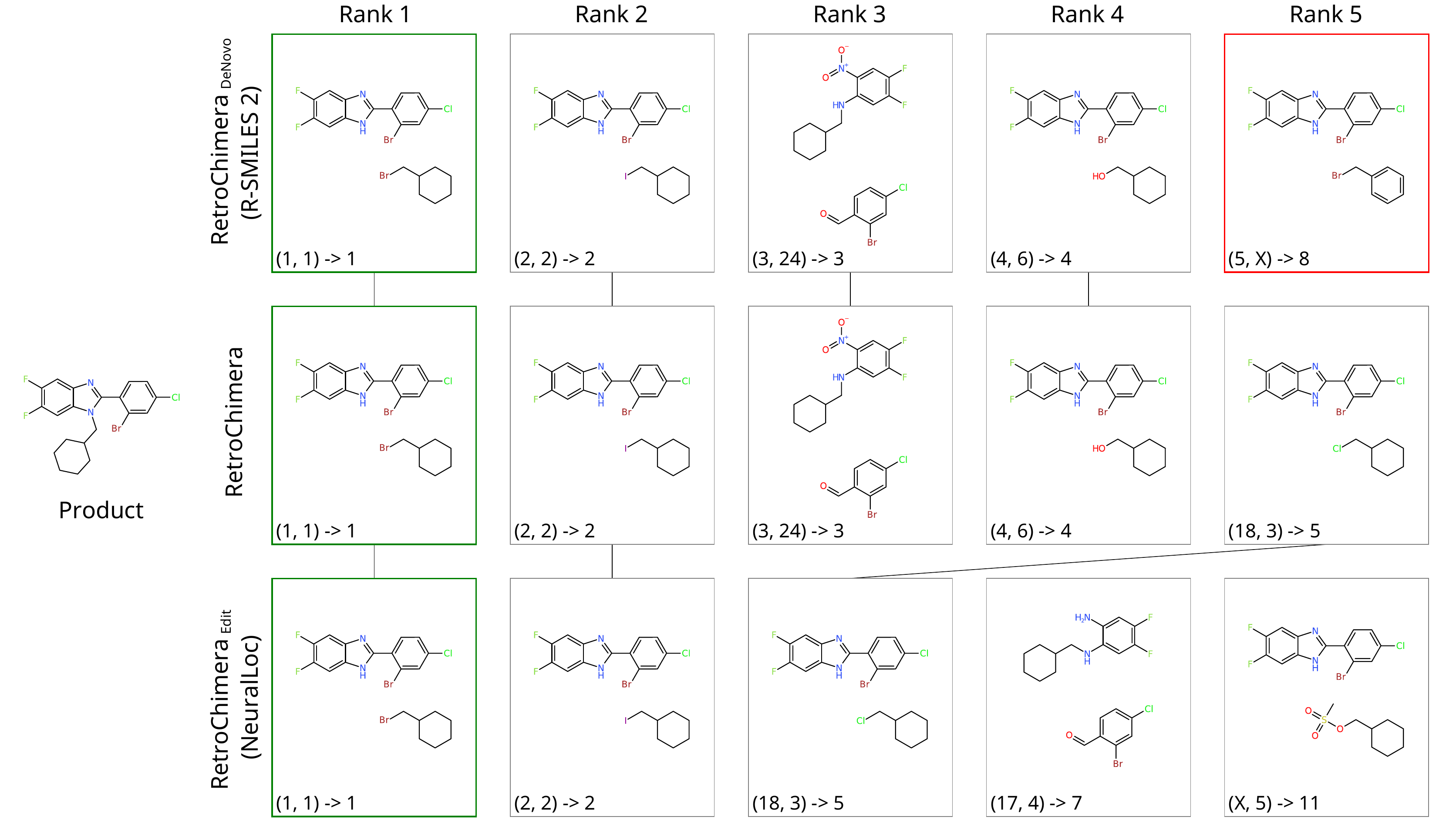}
    \caption{Visualization of how predictions from \smilesmodel~and \locmodel~are combined by \combinedmodel.
    Molecule in row $i$ and column $j$ is the $j$-th reactant set predicted by the $i$-th model.
    \texttt{(A, B) $\rightarrow$ C} denotes that a prediction was rank \texttt{A} in the output of \smilesmodel, rank \texttt{B} in the output of \locmodel, and rank \texttt{C} in the combined output (\texttt{X} signifies a prediction was not found in one of the lists).
    Segments connect molecules that are shared.
    Green box is ground-truth, red box highlights a hallucinated prediction which is chemically implausible.}
    \label{fig:ensembling_visualization}
\end{figure*}

\paragraph{SimpRetro}

To benchmark \combinedmodel~in multi-step search we integrate our models into syntheseus, and start with an initial exploration of success rate on a dataset collected by Li et al.~\cite{li2024challenging}
We reuse the experimental setup from SimpRetro, including the choice of the search algorithm, building blocks (23.1M commercially available molecules from \textit{eMolecules}), GPU type, and time limit.
We consistently see higher success rates than SimpRetro, with \combinedmodel~also outperforming its constituents, and obtaining close to 100\% solve rate under the largest time limit~(Figure~\ref{fig:search_solve_rate}).
However, the creation of the SimpRetro test set did not control for similarity to Pistachio training data.
To supplement this analysis, we move to a dataset of targets based on Pistachio.

\paragraph{Pistachio}

To collect a challenging search dataset sufficiently distinct from training data, we used Pistachio test products that had high SAScore~\cite{ertl2009estimation} and could not be easily solved through retrosynthetic search with NeuralSym, and selected a diverse subset based on fingerprint similarity (see Methods for details).
This procedure left us with $951$ hard targets which we split into $151$ for validation and $800$ for testing.

We search with Retro*~\cite{chen2020retro} using the same building block set as in SimpRetro.
To ensure a fair comparison, we first tuned temperature for every model on validation targets, and then used the best value for test targets.
Generally, all of our models yield a better solve rate than baselines, with \locmodel~performing best early on due to its higher efficiency, but losing to \smilesmodel~and \combinedmodel~in the long run (Figure~\ref{fig:search_solve_rate}).
\combinedmodel~performs best for medium-to-long search times, and finds routes for even highly challenging molecules (Figure~\ref{fig:exampleroutes}).

\section*{Qualitative analysis}

In order to understand the complementary strengths of our proposed models, as well as how ensembling manages to improve upon their results, we run qualitative analyses using the models trained on Pistachio.

\paragraph{Quality and alignment assessment by experts}
To measure the quality of model predictions, we conducted double-blind AB-tests comparing pairs of models or a single model with dataset ground-truth. Here, predictions for the same target from two sources were presented to PhD-level organic chemists, who were asked to express preference for one of the options.

After gathering $599$ comparisons from $9$ experts covering various pairs of sources, we grouped based on prediction rank in the corresponding model, and mapped results within each group to Bradley-Terry scores, which we used to estimate the probability of each model beating ground-truth (Figure~\ref{fig:preference}). We find that chemists significantly prefer~\combinedmodel's top prediction over the dataset ($P < 0.05$, mean preference rate $\approx 64\%$);~\combinedmodel~also outperforms its submodels but that does not reach statistical significance.
As a control, we employed a baseline which naively applies uncommon templates without any ranking, and mixed $46$ baseline pairs into $599$ described above; we find that baseline predictions were rejected in over $93\%$ of cases, confirming that raters were staying attentive.
See Methods for details and Extended Data Figure~\ref{fig:preference_raw} for the raw results. This is the first time a model is able to provide predictions that are more aligned to chemists' expectations than the actual reference reactions the model has been trained on.

\begin{figure*}[!ht]
    \centering
    \begin{tabular}{ll}
    \sidesubfloat[]
    {\includegraphics[scale=0.35]{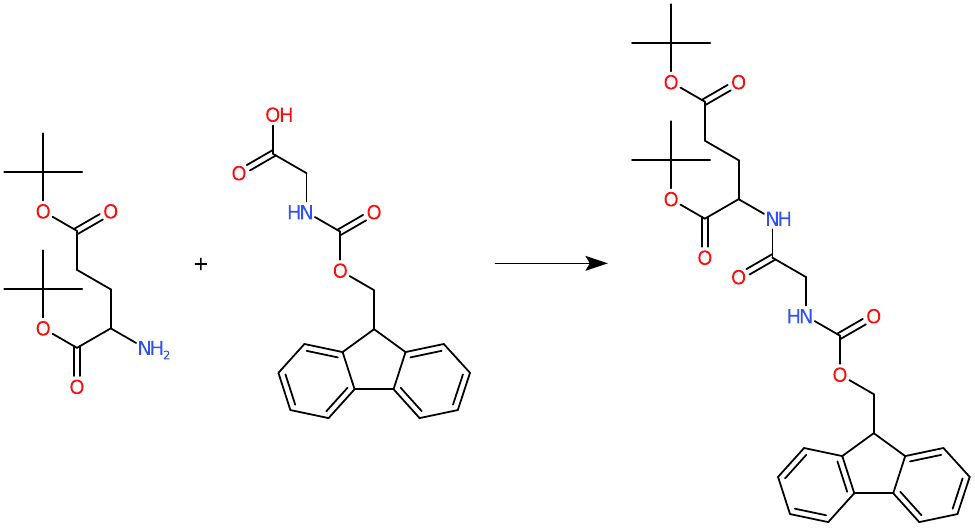}
    } &
    \sidesubfloat[]{\includegraphics[scale=0.35]{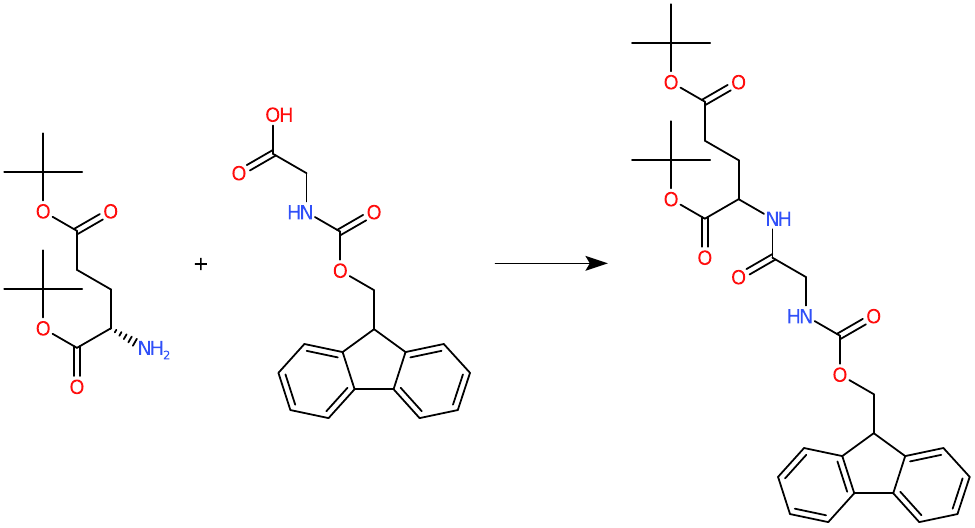}}\\
    \sidesubfloat[]{\includegraphics[scale=0.35]{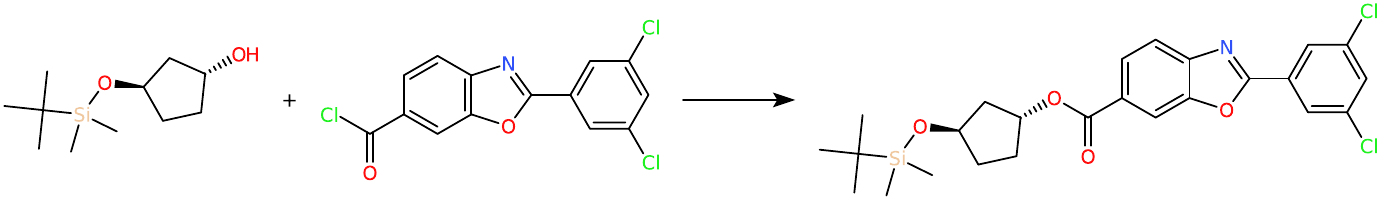}} &
    \sidesubfloat[]
    {\includegraphics[scale=0.35]{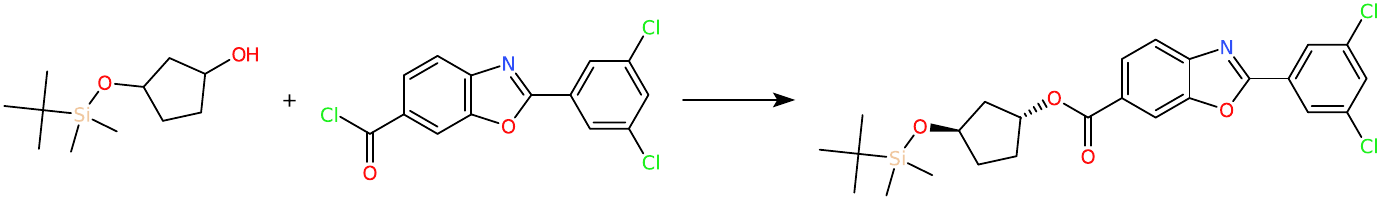}}\\
    \sidesubfloat[]{\includegraphics[scale=0.35]{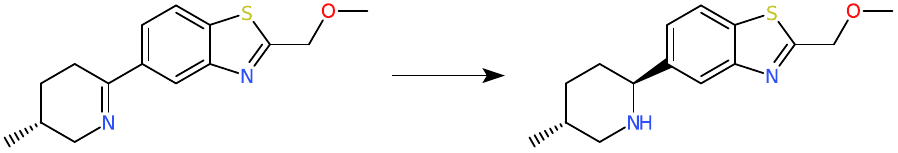}} &
    \sidesubfloat[]{\includegraphics[scale=0.35]{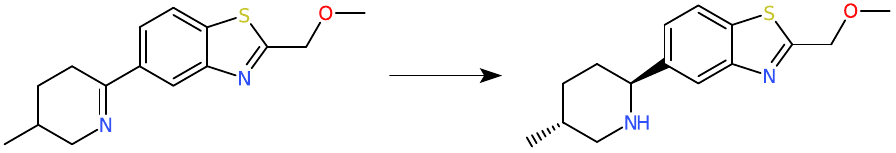}} \\
    \sidesubfloat[]{\includegraphics[scale=0.35]{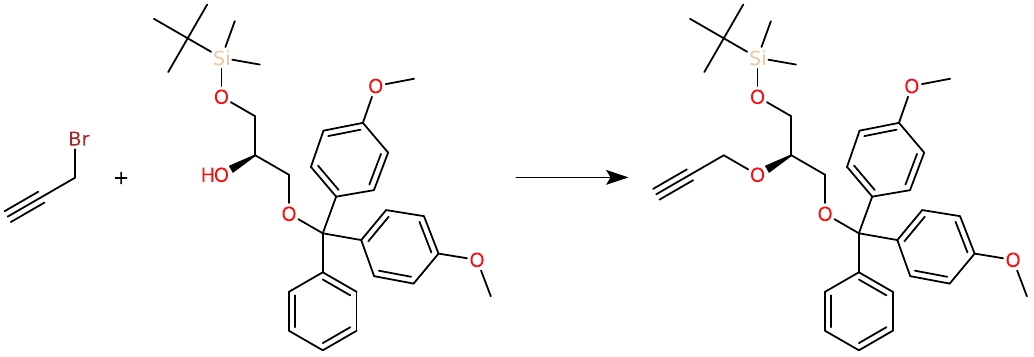}} &
    \sidesubfloat[]{\includegraphics[scale=0.35]{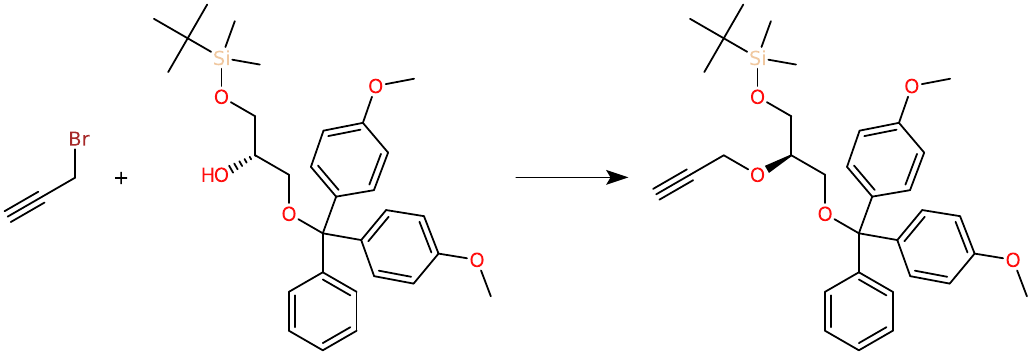}} \\
    \sidesubfloat[]{\includegraphics[scale=0.35]{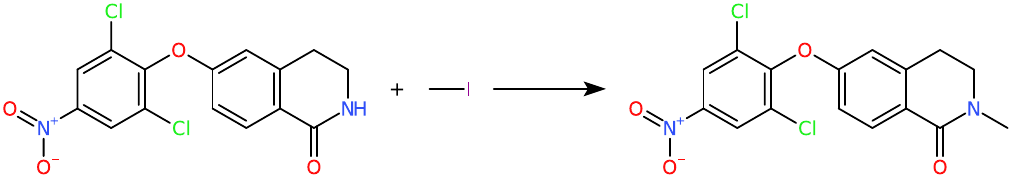}} &
    \sidesubfloat[]{\includegraphics[scale=0.35]{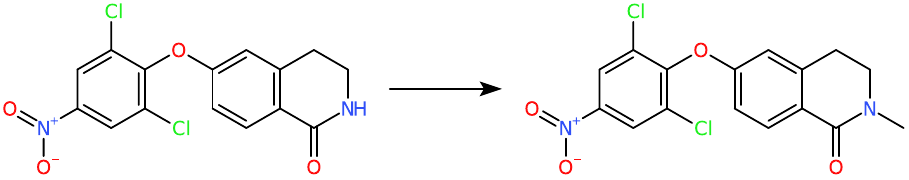}} \\
    \end{tabular}
    \caption{Examples of the denoising behaviour of \combinedmodel. Left column (\textbf{a}, \textbf{c}, \textbf{e}, \textbf{g}, \textbf{i}) are model predictions, in each case preferred by expert chemists. Ground truth examples (right column; \textbf{b}, \textbf{d}, \textbf{f}, \textbf{h}, \textbf{j}) are taken from test set. Examples \textbf{a}-\textbf{h} demonstrate denoising for stereochemistry assignments. Our model has learned to ignore likely incorrect assignments (right) and instead is aligned with expert expectations (left). Furthermore, the model exhibits the ability to infer missing reactants. In example \textbf{j} the test data does not specify the exact alkylating agent, whereas the model infers to use methyl iodide (Example \textbf{i}).}
    \label{fig:denoising}
\end{figure*}

\paragraph{Ensembling visualization}

To visualize what errors are being made by our models and how ensembling helps to mitigate them, we used an early version of the feasibility model to mine unlikely predictions on Pistachio test data.
In a selected example (Figure~\ref{fig:ensembling_visualization}) all models correctly predict the ground-truth as their top prediction, but diverge further down the list, where the 5th output from \smilesmodel~is an erroneous version of the ground truth with one of the rings turned aromatic.
As this is chemically implausible and not covered by templates, it is not predicted by \locmodel, and thus downweighed in \combinedmodel's outputs in favour of predictions shared by the submodels.
The unlikely prediction still appears in \combinedmodel's output; while in this case it may seem undesirable, many predictions made only by \smilesmodel~are correct, which is reflected in the ensembling weights.
Our ensembling formalism permits a solution in which all outputs shared by both models are ranked above those predicted by one, but empirically this is suboptimal.

\begin{figure*}[h]
    \includegraphics[width=0.88\linewidth]{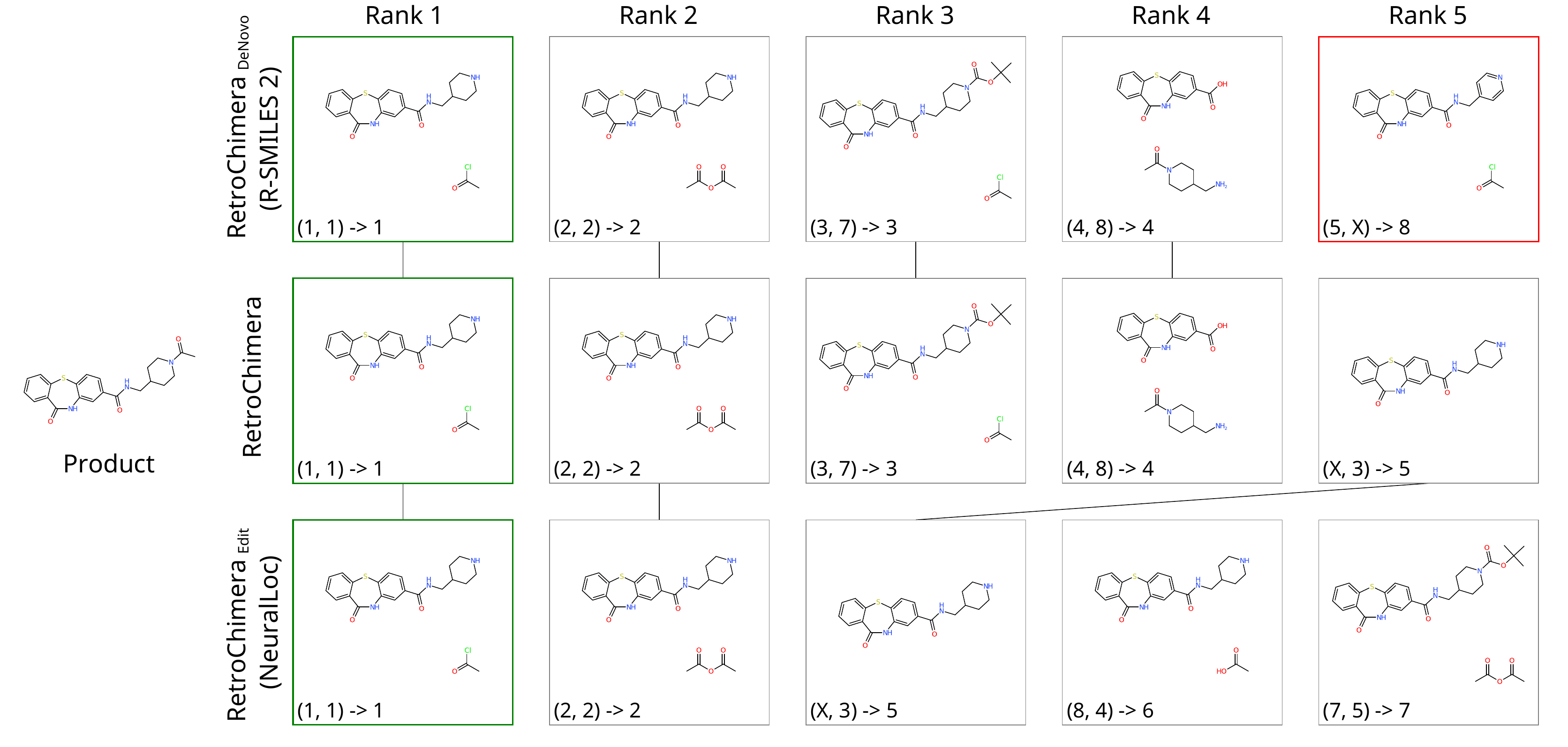}\vspace{0.4cm}\\
    \includegraphics[width=0.88\linewidth]{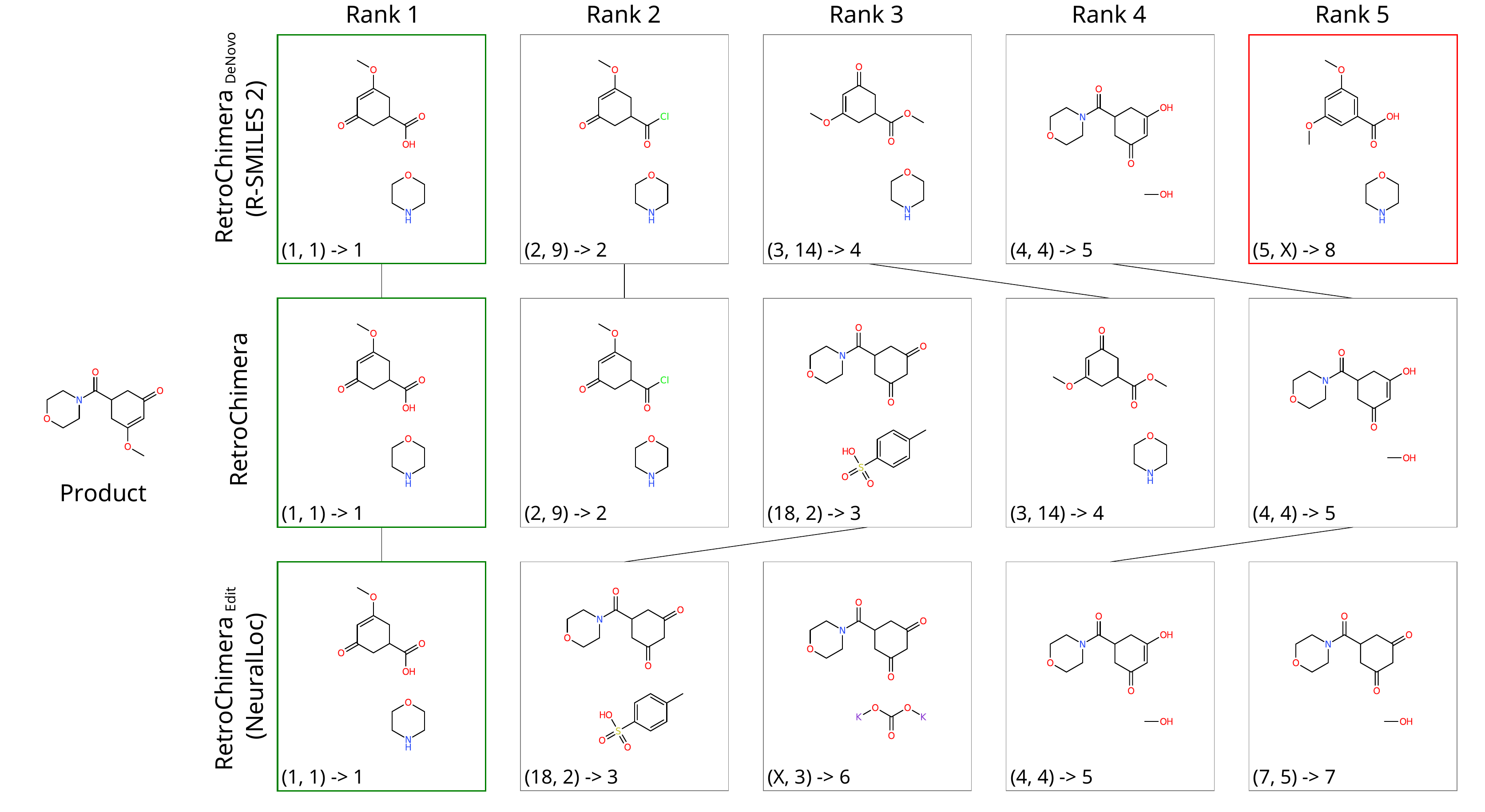}
    \caption{Extended examples of ensembling improving over individual models. Similarly to Figure~\ref{fig:ensembling_visualization}, we see \smilesmodel~can fail to correctly reproduce the right bond pattern in a ring copied from the input product.}\label{fig:pistachio_analysis_extended_1}
\end{figure*}

\begin{figure*}[h]
    \includegraphics[width=0.88\linewidth]{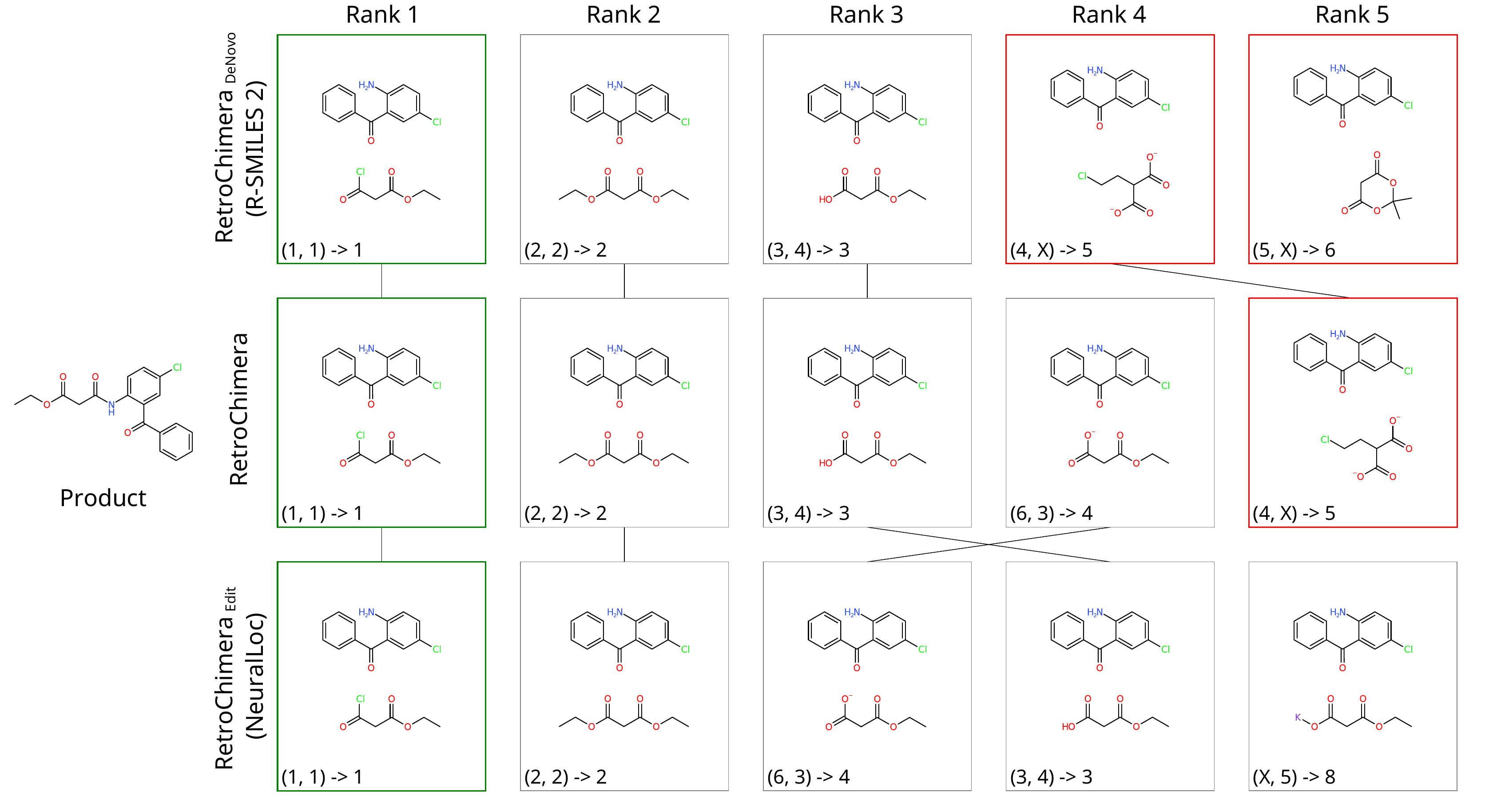}\vspace{0.4cm}\\
    \includegraphics[width=0.88\linewidth]{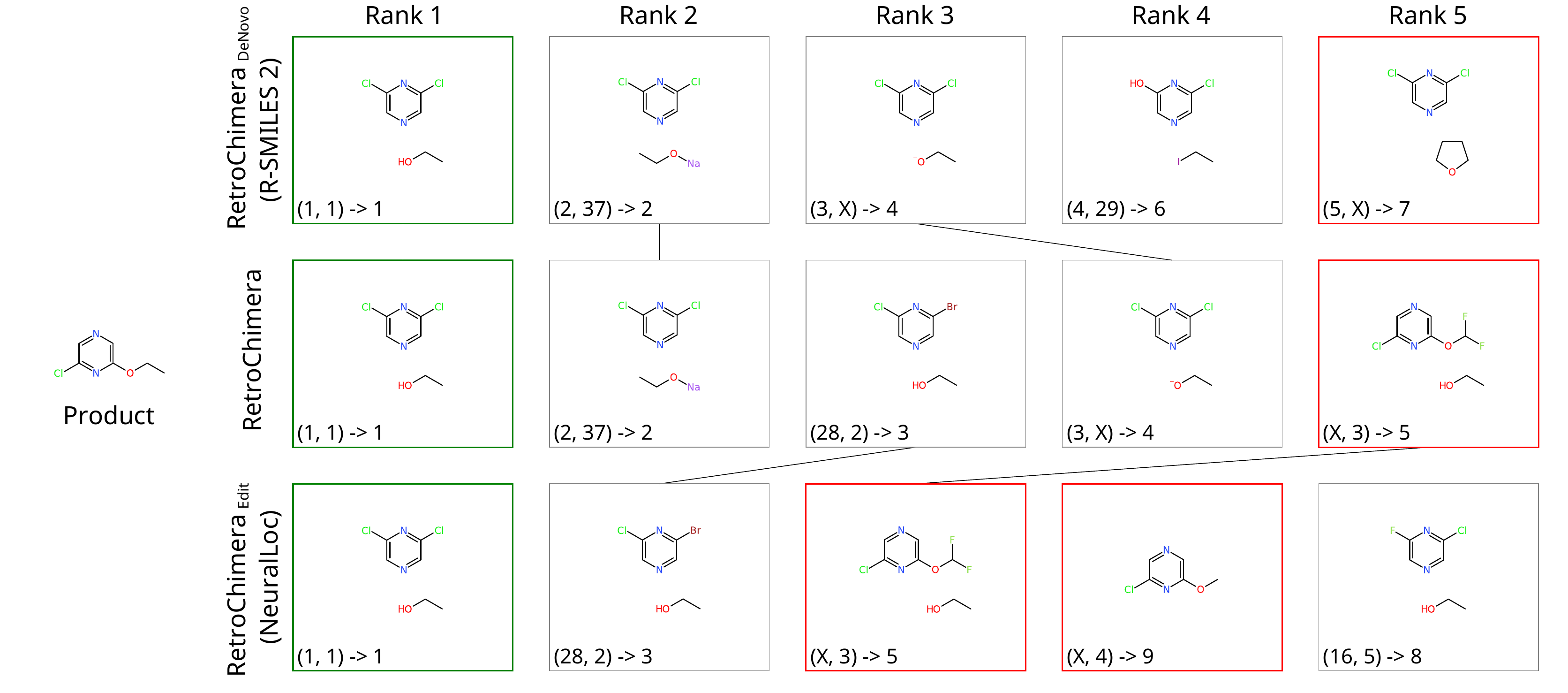}
    \caption{For certain inputs, the \smilesmodel~model might predict bond-breaking reactions which
are chemically implausible (ranks 4 and 5 in the top example; rank 5 in the bottom one). These cases are downweighed during model ensembling as they are not predicted by \locmodel. In contrast, \locmodel~can fail due to noise in the underlying data and incorrect template extraction (ranks 3 and 4 in the bottom example), which is in turn down-ranked by \smilesmodel, highlighting the power of the ensembling approach.}
    \label{fig:pistachio_analysis_extended_2}
\end{figure*}

\begin{figure*}[h]
    \includegraphics[width=0.88\linewidth]{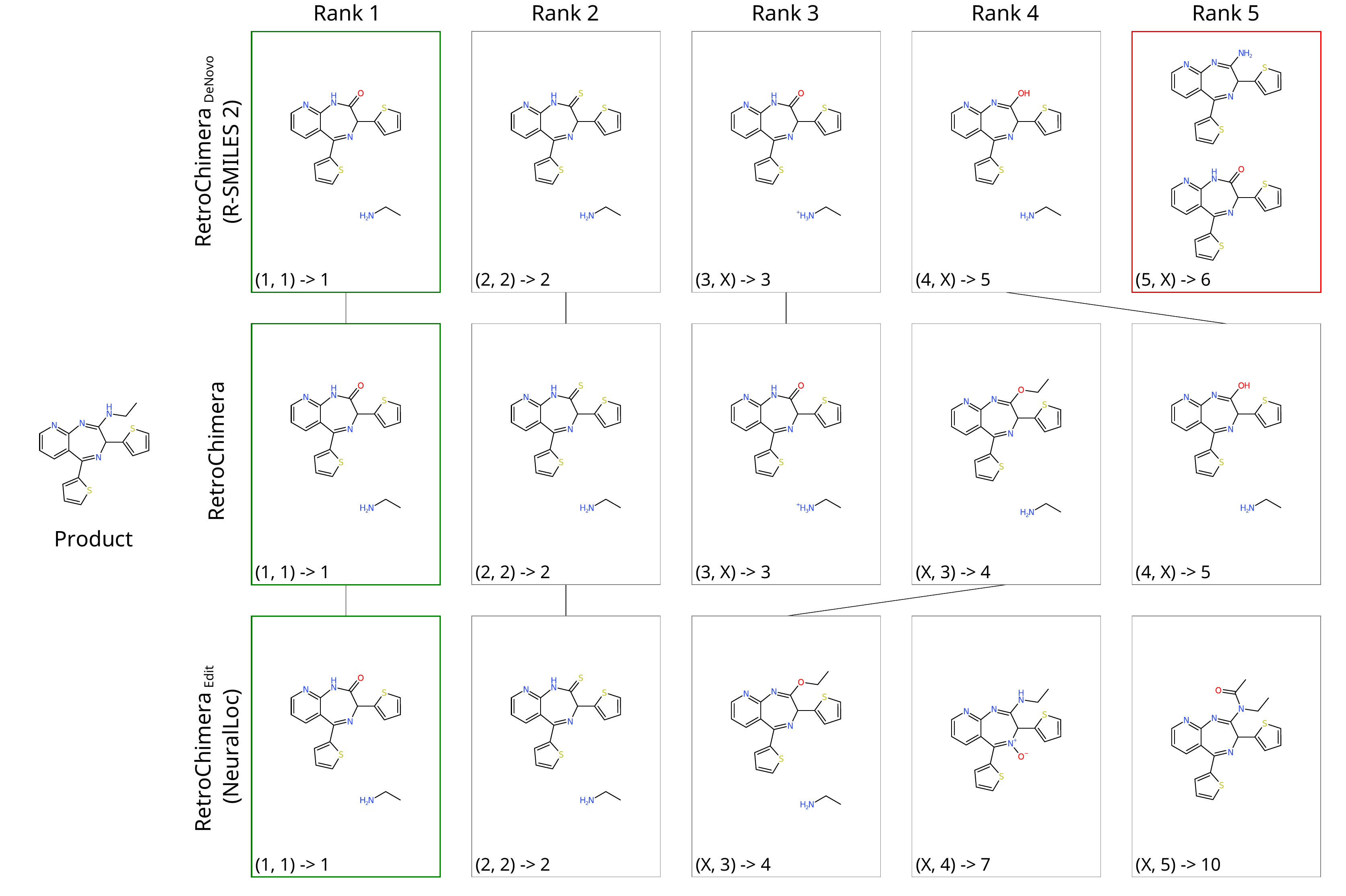}\vspace{0.4cm}\\
    \includegraphics[width=0.88\linewidth]{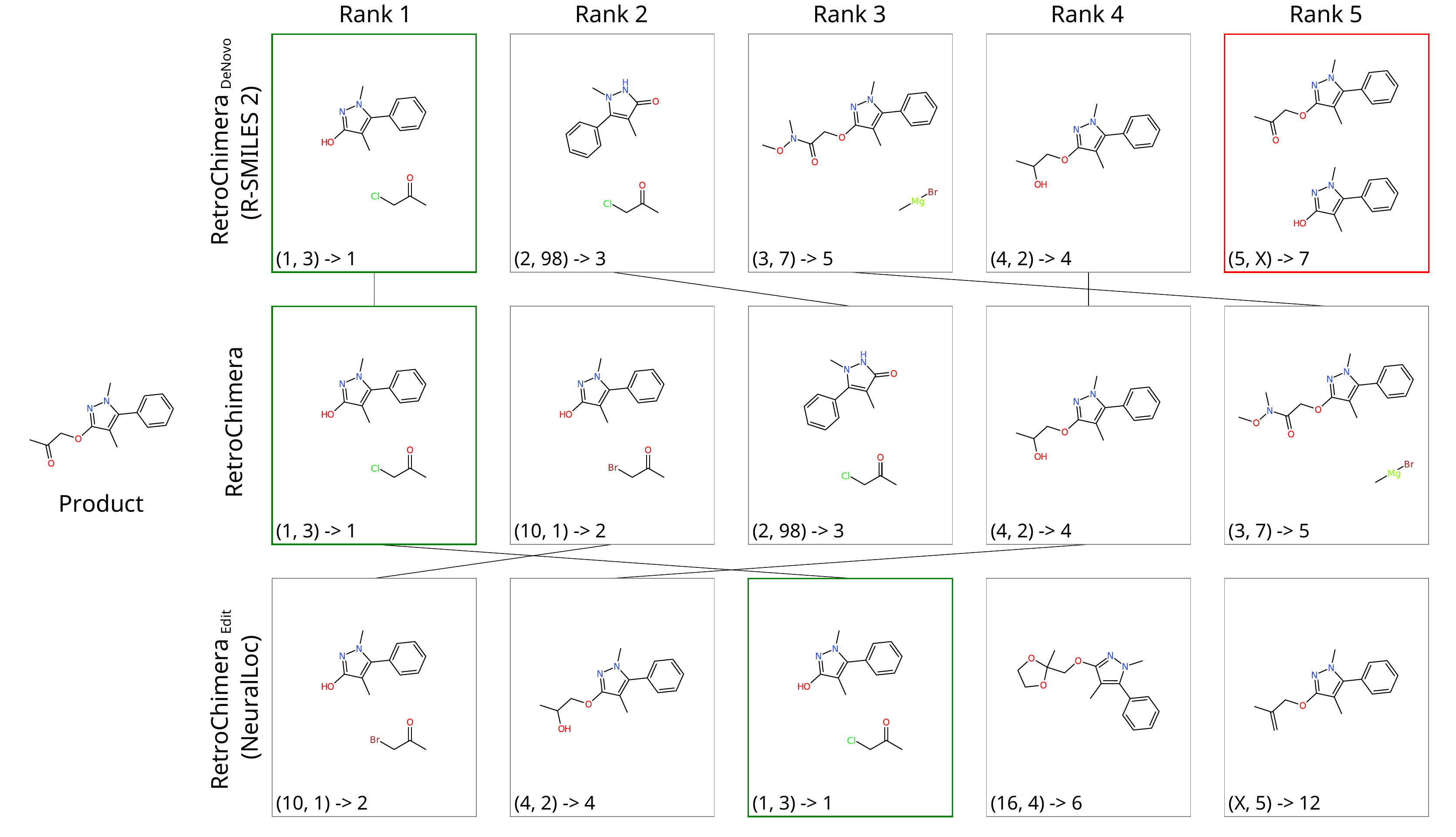}
    \caption{In certain cases, the \smilesmodel~model appears to produce the same reactant twice, either as an exact copy or with minor variation.}
    \label{fig:pistachio_analysis_extended_3}
\end{figure*}

Furthermore, we found cases of copy errors (Extended Data Figure~\ref{fig:pistachio_analysis_extended_1}), implausible bond-breaking reactions (Extended Data Figure~\ref{fig:pistachio_analysis_extended_2}), and duplicating one of the reactants (Extended Data Figure~\ref{fig:pistachio_analysis_extended_3}); all produced by one submodel and downranked in \combinedmodel.
However, these erroneous examples were highly cherry-picked, representing a tiny minority of all predictions.

\paragraph{Denoising of potentially erroneous data}
In past work, AI retrosynthesis models have been criticized for their potential inability to deal with noise in the training data, which is expected to be present in large reaction datasets.\cite{mikulak2020computational} 
However, correctly trained probabilistic models, such as the ensemble component models in this work, can become robust towards errors in the training set, and effectively denoise the data. 
For this purpose, we qualitatively inspected random test reactions for given products from the Pistachio dataset that our model was unable to recover in its top 50 predictions, and asked expert chemists for an assessment (Figure~\ref{fig:denoising}). We found that representative erroneous ground truth test reactions for example contain stereochemistry or other assignment errors, while our model returns the reactions that the chemist would expect, highlighting the ability of the model to deal robustly with partially noisy data and align with scientists' expectations.

\section*{Limitations}
Despite increased robustness, ML-based retrosynthesis models are not free from hallucinated outputs, especially far away from the training distribution. This can partially be mitigated by combining retrosynthesis models in a pipeline with reaction feasibility and forward prediction models, as demonstrated in prior work.\cite{segler2018planning, coley2019robotic} Another limitation stems from systematic errors in the training data, which can be mitigated by improved data curation.

\section*{Conclusion}

In this work, we introduced a framework for building powerful retrosynthesis models by ensembling. Instantiated with two new models with different inductive biases, each exhibiting favorable performance in their own categories, we introduced \combinedmodel, and demonstrated its efficacy on commonly used datasets, providing key insight into the strengths of different model classes. For the first time, we have demonstrated close to optimal retrieval for rare reaction classes, thus allowing retrosynthesis models to essentially become soft reaction databases, and shown that the ensemble is preferred by expert organic chemists in terms of quality.
In experiments on both existing and new benchmarks, we validated that \combinedmodel's strong performance carries over to multi-step search.

Our results open up ensembling strategies as a new dimension to improve retrosynthesis models, and demonstrate that deep learning method development,  leveraging latest progress in Transformers and powerful representation learning for chemical transformations, continues to be a fruitful path to improving model performance. 
With the parallel development of increasing availability of standardized, high-quality high throughput experimentation data, we anticipate further acceleration towards the goal of fully closed-loop synthesis planning, orchestration and execution.

\section*{Code and Model Availability}
\combinedmodel~is available through Azure Foundry at \href{https://ai.azure.com/catalog/models/RetroChimera}{ai.azure.com/catalog/models/RetroChimera}.
Code is currently being prepared for release on GitHub.

\section*{Acknowledgments}

We would like to thank our anonymous industrial chemists, recruited in Seattle, Beijing, Cambridge, Basel and at the Bürgenstock conference, for providing their assessments. We thank Bichlien Nguyen, Elise van der Pol, Fabio Lima, Jake Smith, Jose Garrido Torres, Paola Gori Giorgi, Rianne van den Berg and Tao Qin for discussions and feedback; Jingyun Bai for design; Usman Munir and Shoko Ueda for project management; Hannes Schulz, Jean Helie, Maik Riechert and Ran Bi for engineering support. 
Moreover, we would like to acknowledge the whole AI for Science team at Microsoft Research, as well as collaborators at Novartis, for their ongoing support.

\section*{Contributions}
Krzysztof, Guoqing and Marwin conceptualized the work. Guoqing created \smilesmodel, while Krzysztof developed \locmodel~and the ensembling procedure. Hubert and Aleksei performed evaluations on Novartis data. Holger, Rishi and Mike advised the project. Piotr built the feasibility model, which was scaled up to Pistachio by Krzysztof.
Guoqing built the forward model.
Marwin performed data curation, analysed model outputs (with help from Junren), and conducted the study with expert chemists.
Austin built the backbone for multi-step search, which was extended by Krzysztof who ran the search experiments.
Paper was written by Krzysztof, Guoqing and Marwin, with feedback from the other authors. The project was mentored by Marwin.

\bibliographystyle{naturemag}
\bibliography{main}

\clearpage

\section*{Methods}

\subsection*{Prior Work on Ensembling}
While ensembling for reaction prediction and retrosynthesis has been attempted, results have been limited so far. Schwaller et al.~\cite{schwaller2019molecular} ensemble up to 20 forward models, but report only minimal gains at the cost of significantly slower inference. 
However, they employ the default method in OpenNMT~\cite{klein-etal-2017-opennmt}, which averages next token probability distributions predicted by the different models, and is limited to models sharing the same output space. 

Combinations of models have been reported with specialized models for ring-forming reactions\cite{thakkar2020ring} or enzymatic catalysis\cite{levin2022merging,kim2024readretro}. Lin et al.~\cite{lin2022improving} combine outputs from different models, but determining the final order relies on a separately trained ranking model, discarding the rich information present in the order predicted by the original models. Torren-Peraire observed differences in the solutions different single-step models find.\cite{torren2024models} 
In a recent paper by Saigiridharan et al., it was explicitly pointed out that while different models have been combined ad-hoc\cite{torren2024models}, no principled ensembling approach is available.\cite{saigiridharan2024aizynthfinder}

\subsection*{Ensembling}

We learn ensembling parameters $\theta$ using the Adam optimizer~\cite{kingma2014adam} to minimize $\mathcal{L}_{rank} + w_{reg} \cdot \mathcal{L}_{reg}$, where $\mathcal{L}_{reg}$ is a regularization term to ensure relative model importance does not change too rapidly across ranks

\begin{equation*}
\mathcal{L}_{reg} = \frac{1}{m(m-1)} \sum_{i \neq j} 
\frac{1}{k_{max} - 1} \sum_{k = 1}^{k_{max} - 1} \left| \frac{\theta_{i,k}}{\theta_{j,k}} - \frac{\theta_{i,k + 1}}{\theta_{j,k + 1}} \right|
\end{equation*}

We find that a regularization of this form gives a modest improvement for $m = 2$ and is roughly neutral for large $m$; we thus use a small weight of $w_{reg} = 0.2$.

Due to correlations between the rankings produced by the different models, in the majority of cases the relative ordering of $r^+$ and $r^-$ is preserved across all models, especially when $m$ is small. Those cases, while contributing non-zero gradient to $\mathcal{L}_{rank}$ for $T > 0$, are bound to be ranked in the same way for any row-wise decreasing $\theta$. Thus, in practice we skip those pairs $(r^+, r^-)$ in Equation~\ref{eq:lrank} to reduce variance.

\paragraph{Constraining $\theta$}

One could minimize $\mathcal{L}_{rank}$ directly, but small validation set size and poor coverage of cases where $r^+$ appears at higher ranks lead to overfitting and poor generalization.
To fix this, we constrain each $\theta_i$ to be decreasing and convex ($\theta_{i, k} > \theta_{i, k+1}$ and $\theta_{i, k} - \theta_{i, k + 1} > \theta_{i, k + 1} - \theta_{i, k + 2}$), expressing the intuition that lower ranks are less likely to be correct, and differences between ranks are more pronounced closer to the top. Formally, we parameterize $\theta_i$ as $\texttt{flip}(\texttt{cumsum}(\texttt{cumsum}(\texttt{exp}(x_i)))$, where $x_i \in \mathbb{R}^{k_{max}}$ are free parameters, $\texttt{cumsum}$ computes a cumulative sum, and $\texttt{flip}$ reverses the vector.

\paragraph{Implementation details}

To optimize $\theta$, we first map the entire validation set into a single tensor containing ranks of $r^+$ and $r^-$ across all models, which allows $\mathcal{L}_{rank}$ to be computed efficiently through a handful of PyTorch~\cite{paszke2019pytorch} primitives. We do not use batching, and instead optimize the full loss directly for $1000$ steps. Both the learning rate and the temperature $T$ start at $0.1$ and decay by a factor of $0.9$ every $25$ steps. We set the margin $\epsilon$ in Equation~\ref{eq:lrank} to $10^{-4}$.

\subsection*{Fingerprint similarity}

We make use of fingerprint similarity in several aspects of our work: filtering out near matches when constructing the Pistachio test set, bucketing the test samples for Figure~\ref{fig:accuracy_vs_similarity}, and generating synthetic negative reactions for training the feasibility model.

In all cases we use count-based Morgan fingerprints with radius $2$ folded modulo a large prime.
To compute similarity between $x$ and $y$ we employ Tanimoto similarity adapted to count fingerprints~\cite{willett1998chemical, steffen2009comparison}:

\begin{equation*}
\texttt{sim}(x, y) = \frac{\sum_i x_i y_i}{\sum_i x_i^2 + \sum_i y_i^2 - \sum_i x_i y_i}
\end{equation*}

In practice we care about all-pairs similarities between two large sets of molecules; we thus make use of an efficient GPU-based implementation that pads the fingerprints to the nearest power of $2$ and rephrases computing $\texttt{sim}$ in terms of matrix multiplication.

\subsection*{Editing submodel (\locmodel)}

\paragraph{Input featurization}

To featurize the input product, we follow prior work~\cite{chen2021deep} and represent a molecule as a graph $\mathcal{G} = (V, E)$ with nodes $V$ and edges $E$ corresponding to atoms and bonds, respectively. To construct domain-specific node and edge features, we employ the featurizers available in the dgllife library~\cite{dgllife}. Specifically, we use \texttt{WeaveAtomFeaturizer} for atoms and \texttt{CanonicalBondFeaturizer} for bonds. Following LocalRetro~\cite{chen2021deep} we set the atom types supported by the atom featurizer to \texttt{dgllife.data.uspto.atom\_types} extended by Tantalum. We do not include loops in $\mathcal{G}$ by setting \texttt{self\_loop=False}.

\paragraph{Template extraction}
Templates were extracted with rdchiral.\cite{coley2019rdchiral} We note that alternative approaches for template extraction\cite{christ2012mining, heid2021influence}, minimal templates\cite{segler2018planning, chen2021deep}, or manually coded rules\cite{segler2017neural} in combination with template prediction have been described in prior work and could potentially lead to improved results in future work. 

\paragraph{Template featurization}

Prior work has explored simple template featurization by converting both sides to molecular fingerprints~\cite{seidl2022improving}.
This offers limited flexibility, and only produces aggregate representations, while \locmodel~requires node embeddings to perform localization; we therefore design a new template featurization method to meet this desiderata, which turns an input template into a graph.

As both sides of the template resemble molecular structures, a starting point is to convert them into two graphs $\mathcal{G}_L = (V_L, E_L)$ and $\mathcal{G}_R = (V_R, E_R)$, respectively.
Structures involved in templates are often not fully complete or valid molecules, thus it is not possible to reuse the input featurizer directly.
However, we find that if we switch to a basic atom featurizer (\texttt{CanonicalAtomFeaturizer} without the chiral tag feature), it is enough to parse the molecules using \texttt{MolToSmarts} followed by calling \texttt{UpdatePropertyCache(strict=False)} to get the graph featurization to work successfully.
Apart from standard features that are taken into account by the atom featurizer, an atom on the left-hand side of a template can also be associated with an atom SMARTS -- a logical pattern describing more nuanced match conditions. In principle, these patterns could be parsed and encoded via a specialized procedure invariant to equivalent logical transformations; for simplicity, we instead opt for a simple one-hot encoding over a vocabulary of atom SMARTS patterns that occur in the data. Next, we add binary features distinguishing $V_L$ from $V_R$ to encode directionality. The last ingredient is to relate $\mathcal{G}_L$ to $\mathcal{G}_R$ by converting the atom mapping to a set of edges $M = \{(u, v): u \in V_L, v \in V_R, u \text{ is matched to } v\}$; these edges are assigned a special edge feature to clearly differentiate from $E_L \cup E_R$.
We define the graph representing the entire template as $\mathcal{G} = (V_L \cup V_R, E_L \cup E_R \cup M)$.

We note that our template featurization procedure is invariant under certain operations that do not affect the semantics of the template, including varying the linearization of the graphs, and permuting the atom mapping identifiers. Two syntactically different representations of the same template will therefore be mapped to the same graph, which can serve a similar purpose to template canonicalization algorithms~\cite{heid2021influence}.

\paragraph{Architecture}

Bulk of the neural processing in \locmodel~is performed by two separate GNNs, $\text{GNN}^{\text{in}}$ and $\text{GNN}^{\text{tpl}}$, which -- after several message passing layers interleaved with normalization and dropout -- produce atom representations $h^{\text{in}}_v$ and $h^{\text{tpl}}_v$, respectively for atoms in the input product and the template.
Both GNNs have a similar architecture based on the PNA~\cite{corso2020principal} message passing scheme as implemented in PyTorch Geometric~\cite{fey2019pyg}.
We experimented with a GPS layer~\cite{rampavsek2022recipe} from Graphium~\cite{beaini2023towards} to extend PNA with global attention, and found it results in a minor performance improvement but significantly higher memory requirement.
This trade-off was only beneficial on the small USPTO-50K dataset, thus we use PNA combined with GPS on USPTO-50K, and only PNA on USPTO-FULL and Pistachio.
As one of the downstream objectives is graph-level, representations $h^{\text{in}}_v$ and $h^{\text{tpl}}_v$ are aggregated similarly to prior work~\cite{maziarz2022learning} using two separate aggregation layers based on multi-head attention to form $h^{\text{in}}$ and $h^{\text{tpl}}$, respectively.
Due to a slight deficiency in the expressivity of our graph-level aggregation method, disconnected templates formed by repeating a fixed component a varying number of times are assigned the same representation, which would prevent the model from differentiating those templates downstream.
Thus, we also introduce an additional template embedding of size $d_{\text{free}}$, which is learned end-to-end as opposed to being produced by the template encoder, and concatenate that to $h^{\text{tpl}}$. 
Finally, we linearly project graph-level representations of both input and template into a shared dimension $d_{\text{clf}}$; those projections are then used for the classification objective.
Network sizes vary across datasets, and were informed by overfitting concerns on USPTO-50K, and memory considerations on larger datasets (see Extended Data Table~\ref{tab:locmodel-arch-hparams} for the exact values).

\paragraph{Classification objective}

For classification, the input representation is multiplied by stacked template representations, and the resulting dot products are interpreted as unnormalized template selection scores.
Unlike MHNreact~\cite{seidl2022improving}, our template processing is learned, and thus templates used for classification have to be repeatedly encoded in each batch.
The cost to do so grows with the number of templates and at sufficient scale becomes prohibitive.
While on USPTO-50K we can encode all templates afresh in each forward pass, on USPTO-FULL and Pistachio doing so would require excessive amounts of GPU memory.
Therefore, on larger datasets we only include a subset of templates in the classification objective, which include the ground-truth answers in a given batch and $r^{\text{clf}}$ randomly sampled templates per batch input as additional negatives; those negatives participate in classification for all inputs, not only those they were sampled for.
While we use a simple softmax cross-entropy classification loss for the case of including all templates in each forward pass, when including a subset we found that the losses stemming from different templates have to be re-weighted according to template frequency to allow for learning appropriate marginals.
In this case we use a sigmoid pairwise classification loss inspired by prior work~\cite{zhai2023sigmoid}.
We found increasing $r^{\text{clf}}$ generally tends to improve results, and so we set it as high as possible given memory constraints (Extended Data Table~\ref{tab:locmodel-arch-hparams}).

\paragraph{Localization objective}

Localization requires assigning each atom in the left-hand side of the template ($V_L$) an appropriate atom in the input ($V$).
To that end, we multiply $h_v^{\text{tpl}}$ for $v \in V_L$ with $h_u^{\text{in}}$ for $u \in V$, and interpret resulting dot products as unnormalized localization scores, which are passed through a softmax along the template atoms dimension.
The primary purpose of localization is to differentiate outputs resulting from applying a single template, but during inference we use a combination of classification and localization to rerank all outputs globally; thus it is beneficial for the localization subnetwork to be exposed to other templates beyond the ground-truth one during training.
Therefore, in practice we use not only the node representations extracted for the ground-truth template, but also include $r^{\text{loc}}$ other templates from the current batch that best match a given input according to classification scores; this requires minimal additional computation as node representations for those templates were already computed for classification.
The final localization loss is as a sum of cross-entropy losses over the template nodes.
For nodes in the ground-truth template the target is to select the corresponding atom in $V$, whereas for nodes in additional negative templates the network is trained to instead select an auxiliary $h_{\text{neg}}^{\text{tpl}}$ representation, which is concatenated to $h_v^{\text{in}}$ and trained end-to-end.
Often there may be several localizations of the ground-truth template that result in correct predicted reactants; we label all of those localizations during preprocessing, so that the loss for atoms in the ground-truth template can use a uniform distribution over all correct choices in $V$ as the target.

\paragraph{Training}

We train \locmodel~by minimizing a sum of the classification and localization losses.
Training proceeds for a fixed number of epochs followed by checkpoint selection according to validation MRR.
Following prior work~\cite{zhong2022root} we select several best checkpoints (typically 5-10), and perform checkpoint averaging in parameter space to produce the final weights.

\paragraph{Inference}

During training, atom- and graph-level template representations evolve with each update to $\text{GNN}^{\text{tpl}}$, and thus have to be recomputed each time they are used downstream.
However, upon saving each checkpoint we encode all templates in the library and include the resulting outputs in the checkpoint file; this allows for fast inference as $\text{GNN}^{\text{tpl}}$ no longer needs to be used.
Given a test input, we first multiply $h_{\text{in}}$ with template representations and extract $r^{\text{app}} \cdot n$ top-scoring templates to apply, where $n$ is the number of results requested downstream; this step is identical to performing inference in the NeuralSym model.
$r^{\text{app}}$ is set to $1$ during search, and to a larger value for single-step evaluation (Extended Data Table~\ref{tab:locmodel-arch-hparams}).
After applying the selected templates -- which can be done efficiently using multiprocessing -- for each template we group the predictions based on the resulting reactants, in order to account for several localizations producing the same result.
Next, we rerank all unique outputs according to $s^{\text{clf}} + w^{\text{loc}} \cdot s^{\text{loc}}$, where $s^{\text{clf}}$ is the normalized template log-probability, $s^{\text{loc}}$ is the average normalized localization log-probability over template atoms, and $w^{\text{loc}} = 2.25$ is a coefficient chosen empirically.
When computing $s^{\text{loc}}$ we sum localization probabilities over potentially several correct choices, as highlighted by the aforementioned grouping.
Finally, we truncate the output list to $n$ results ($100$ for single-step benchmarking, $50$ during search).

\subsection*{De-Novo submodel (\smilesmodel)}

\paragraph{Architecture}

We build upon R-SMILES~\cite{zhong2022root}, and train an encoder-decoder model based on a Transformer backbone~\cite{vaswani2017attention}.
Unlike previous work~\cite{klein-etal-2017-opennmt} we reimplement the model from scratch using PyTorch~\cite{paszke2019pytorch}, 
allowing us to freely customize the architecture.
We applied key modifications described in the main text, which were inspired by the recent success of large language models such as Llama~\cite{touvron2023llama2openfoundation} and Mistral~\cite{jiang2023mistral7b}.

\paragraph{Data augmentation} 
Previous studies~\cite{zhong2022root,han2024retrosynthesis} have shown that the general-purpose SMILES neglects the characteristics of chemical reactions, where the molecular graph topology remains largely unchanged from reactants to products.
To address this, we employ root-aligned SMILES~\cite{zhong2022root}, which ensures an aligned mapping between product and reactant SMILES.
This strict mapping, along with a reduced edit distance, simplifies the task for the transformer, allowing it to focus on learning the chemistry involved in reactions rather than syntax. 
We generate multiple input-output pairs as augmented training data by enumerating different product atoms as the root of SMILES.
We apply 20$\times$ augmentation to the USPTO-50K dataset, 5$\times$ to USPTO-FULL, and 10$\times$ to Pistachio.

\paragraph{Tokenization} We follow Schwaller et al.'s~\cite{schwaller2019molecular} regular expression to tokenize products and reactants SMILES into meaningful tokens. The regular expression is defined as:
\begin{verbatim}  
token_regex = "(\[[^\]]+]|Br?|Cl?|N|O|S|P|F|
I|b|c|n|o|s|p|\(|\)|\.|=|#|-|\+|\\\\|\/|:|~|
@|\?|>|\*|\$|\%[0-9]{2}|[0-9])".
\end{verbatim}  

This pattern accounts for the diverse range of symbols and characters within SMILES strings, including brackets, elemental symbols, numbers, and special characters. Notably, it matches sequences within brackets, elemental symbols (including \texttt{Br}, \texttt{Cl}, \texttt{N}, \texttt{O}, \texttt{S}, \texttt{P}, \texttt{F}, \texttt{I}), lower-case letters (\texttt{b}, \texttt{c}, \texttt{n}, \texttt{o}, \texttt{s}, \texttt{p}), parentheses, dot, other symbols (\texttt{=}, \texttt{\#}, \texttt{-}, \texttt{+}, \texttt{\textbackslash}, \texttt{/}, \texttt{:}, \texttt{\raisebox{0.5ex}{\texttildelow}}, \texttt{@}, \texttt{?}, \texttt{>}, \texttt{*}, \texttt{\$}), and two-digit numbers preceded by a percentage symbol, as well as single-digit numbers.

\paragraph{Training objective} We train \smilesmodel~to minimize a standard cross-entropy loss with respect to the token sequence describing ground-truth reactants.

\paragraph{Inference}
During inference we use beam search to find the top $k$ predicted reactant sequences; however, we tailored the beam search logic to retrosynthesis.
Unlike OpenNMT, which keeps completed sequences until two conditions are met -- the pool size equals the beam size and the top-rated sequence in the beam is lower in quality than all in the pool -- we maintain finished sequences in the beam and end only when each sequence in the beam finishes with the EOS token.

We found that this new design makes the top-$k$ list more reliable and significantly improves accuracy, particularly for $k 
\geq 20$, without visibly increasing inference time.

\subsection*{Quality assessment}

\paragraph{Method}

Analysing quality of $k$ top predictions can be confounded by some models having higher top-$k$ accuracy, while others returning less than $k$ outputs altogether.
To study the quality of non-ground-truth predictions directly, we filter the test products to those where all compared models return at least $k$ outputs and recover the ground-truth answer within that; after removing the ground-truths from the output lists, we obtain $k - 1$ non-ground-truth predictions for each input, which are fed into subsequent analysis.

For the comparison in Figure~\ref{fig:forward_and_feasibility} we set $k = 10$ and filter the Pistachio test set down to $113\,135$ products ($\approx 66.7\%$) according to the aforementioned criteria, with $9$ non-ground-truth predictions associated with each.
We then run both quality assessment models on the ground-truth reactions for those products, and calibrate so that each accepts around $95\%$ of ground-truths; for the forward model this translates to accepting a reaction if its product is within top 2 predicted products given the reactants, while for the feasibility model if the predicted feasibility is above $0.1$.

\paragraph{Forward model}
We utilized the same Pistachio reaction dataset and model architecture as the \smilesmodel\ model for the forward model development.
This involved applying 10× R-SMILES augmentation to the Pistachio data in the forward direction. 
After a training for 10 epochs, 
we used the final checkpoint for quality assessment.
To validate the performance of the forward model, we evaluated the trained model on the USPTO-50K test dataset, resulting in top-1 accuracy of 88.6\%, top-3 accuracy of 97.8\%, and top-50 accuracy of 99.9\%. 
When evaluated on the Pistachio test set, the model achieved top-1 accuracy of 70.76\%, top-3 accuracy of 81.3\%, and top-50 accuracy of 87.3\%.
We deemed this accuracy sufficient for conducting convincing quality assessments.

\paragraph{Feasibility model}

To build our feasibility model, we scaled up the approach from prior work~\cite{gainski2024retrogfn} developed on USPTO-50K to the larger Pistachio dataset.
The feasibility model encodes the reactants and product using two separate GNNs, concatenates their aggregated representations, and predicts a single feasibility probability value.
We train it using a standard cross-entropy loss on a dataset consisting of both positive and negative reactions.
We use the Pistachio training data for the former, while the latter is generated synthetically; we gather approximately 10 negative examples for each positive example, for a total of 32M training data points.

We use two separate sources of negative examples: forward template application and similarity-based replacement.
Both hinge on the assumption that if a reaction $R \rightarrow P$ is observed in the data, then other products $P'$ are not formed, i.e. $R \rightarrow P'$ is a negative example.
For the forward template application we follow prior work~\cite{segler2018planning} and use the same templates as used by \locmodel, but applied in the forward direction to reactants sampled from the training data.
For the similarity-based replacement, given a positive reaction $(R, P)$, we find several similar examples $(R', P')$ maximizing $\texttt{sim}(R, R') + \texttt{sim}(P, P')$ where $\texttt{sim}$ is fingerprint similarity defined previously.
We then use $(R', P)$ as the negative example; intuitively, due to the high similarity between $R$ and $R'$, this gives rise to a sample that is more difficult than if one were to pair reactants and products randomly.

\subsection*{Datasets and baselines}

\paragraph{USPTO-50K}

\begin{figure*}[t]
    \centering
    \sidesubfloat[]{\hspace{-0.5cm}\includegraphics[width=0.496\textwidth]{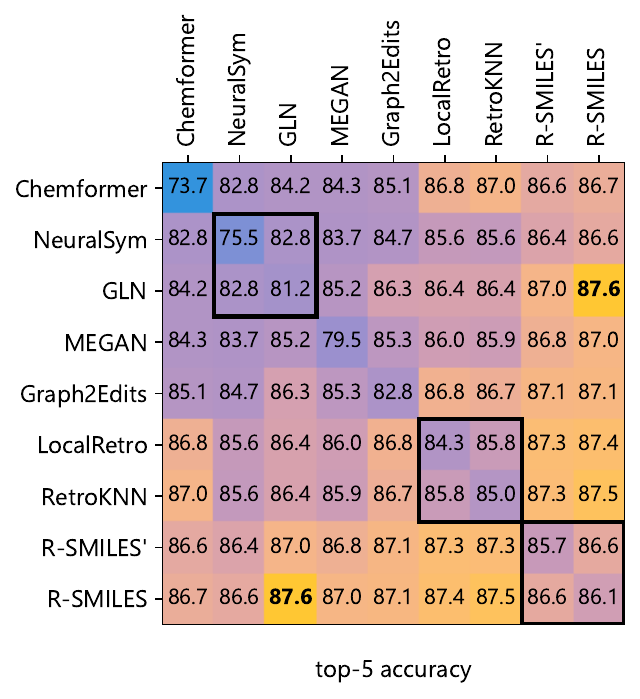}}
    \hfil
    \hspace{0.1cm}
    \sidesubfloat[]{\hspace{-0.5cm}\includegraphics[width=0.496\textwidth]{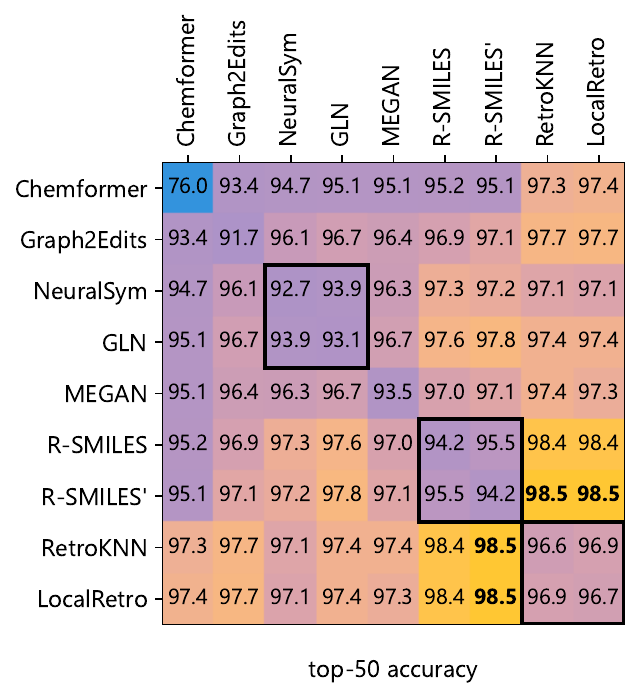}}
    \caption{Top-5 (\textbf{a}) and top-50 (\textbf{b}) accuracy of ensembles of pairs of models. All values are in percent; color palette blue-to-yellow corresponds to low-to-high accuracy (best results shown in bold). 2x2 squares correspond to model clusters which show a limited benefit from being combined: NeuralSym and GLN (both based on standard reaction templates), LocalRetro and RetroKNN (based on minimal templates), and the two checkpoints of R-SMILES. Models are ordered by their result when evaluated in isolation (shown on the main diagonal), with the exception of swapping GLN and MEGAN in the left plot to make the model cluster consecutive. R-SMILES' denotes our retraining of R-SMILES. Off-diagonal entries show ensemble results.}
    \label{fig:ensemble}
\end{figure*}

As baselines for USPTO-50K we selected models integrated into the syntheseus library~\cite{maziarz2023re}, and additionally included our NeuralSym implementation for completeness, and RetroExplainer~\cite{wang2023retrosynthesis} due to strong performance.
We did not include RetroWISE~\cite{zhang2024retrosynthesis} as a baseline, as it utilized extra data from the larger USPTO database. However, it is worth noting that our best ensemble outperforms RetroWISE for $k \geq 5$ despite not using additional data.
We note that some prior works do not compare to R-SMILES on USPTO-50K as the corresponding paper discusses pretraining on USPTO-FULL~\cite{zhong2022root}, but our investigation suggests the checkpoint evaluated in \syntheseus~did not use pretraining, and so it is fully comparable with other USPTO-50K-trained models (this is consistent with the fact that, as seen in Extended Data Figure~\ref{fig:ensemble}, our R-SMILES checkpoint retrained from scratch reached performance close to the released one).

For large ensembles shown in Extended Data Table~\ref{tab:uspto-50k} we included all baseline models from the corresponding table apart from RetroKNN, as its adapter network was trained on $\mathcal{D}_{val}$, which artificially inflates the model's validation result and degrades the performance of ensembles containing RetroKNN.

\paragraph{USPTO-FULL}

Although commonly reported on in prior work, we find many versions of USPTO-FULL are in use, utilizing different methods for filtering and processing; this can be seen through the varying size of the test fold (94696~\cite{wang2021retroprime}, 95389~\cite{yan2020retroxpert}, 95988~\cite{han2024retrosynthesis}, or 96023~\cite{zhong2022root}). Due to this, most reported results on USPTO-FULL are not fully comparable to each other due to using a different test set. For a fair comparison we select a single version of the dataset~\cite{zhong2022root} and only include baselines numbers reported on that version~\cite{chen2021deep,zhong2022root,zhang2024retrosynthesis}, which includes the method with the highest reported top-1 accuracy~\cite{zhang2024retrosynthesis}.
Note that EditRetro~\cite{han2024retrosynthesis} reused the preprocessing script from R-SMILES~\cite{zhong2022root}, but additionally removed 35 test samples with duplicate atom mappings, resulting in a slightly smaller test fold size of 95988 compared to the original 96023.
Since the difference between the two test folds is minimal, we included the values reported by EditRetro in their paper in our table.
Finally, similarly to USPTO-50K, we also included our NeuralSym implementation as a baseline, which we found to produce much stronger performance than reported in prior work.

\paragraph{Pistachio test set}

Time-split validation is considered to be the gold standard for ML model validation in chemistry, as it most closely mimics the prospective use of the models.\cite{sheridan2013time} 
In contrast, random splitting can lead to over-optimistic assessments, especially as reaction data is usually published in clusters, often from the same document (paper or patent), where similar routes are used towards related products.

To construct the time-split test set, we selected reactions added to Pistachio in 2024 as part of the Q2-2024 release.
Based on the Pistachio quality tier assignment we used all reactions from tiers \texttt{S}, \texttt{A}, \texttt{B}; for tiers \texttt{C} and \texttt{D} only reactions with an assigned namerxn name reaction label were used.
All other reactions, including the entire tier \texttt{E}, were rejected.
Finally, we removed reactions of type resolution (RXNO class 11).

We then used fingerprint similarity folded modulo $4093$ to filter out products whose maximum similarity to a training product was at least $0.95$.
Finally, the remaining reactions were processed by the same filtering and deduplication pipeline as the training data.

\paragraph{Bucketing test data}

To produce Figure~\ref{fig:accuracy_vs_similarity}, we bucket Pistachio test data in two ways: based on maximum fingerprint similarity $\texttt{sim}$ to a training product, and based on the frequency of the ground-truth template in the training template library.

Note that \locmodel~only considers templates that appear in training data at least twice, so it is unable to predict a template that occurs once or does not occur at all.
Despite this, as seen in Figure~\ref{fig:accuracy_vs_similarity} (middle), \locmodel~still shows non-zero accuracy on samples with template frequency less than $2$.
This is explained by the fact that several distinct templates could potentially yield the same reactants after being applied to a particular product; hence even if the canonically determined template for a test sample is not available to \locmodel, there may be another template in the library that gives rise to the right reactant set.

\subsection*{Search benchmark}

\paragraph{Target set construction}
To build a challenging test set for search, we started with $146\,393$ Pistachio test products and performed the following steps:
\begin{itemize}
    \item Filter out building blocks ($138\,699$ targets left).
    \item Filter out products whose SAScore is below $4$ ($25\,482$ targets left).
    \item Filter out products containing deuterium atoms ($23\,850$ targets left).
    \item Cluster products with HDBSCAN~\cite{campello2013density} (minimum cluster size $3$, cluster merge threshold $0.15$) using fingerprint similarity $\texttt{sim}$ to define a distance measure. Discard $4437$ noisy (unclustered) products, and pick the highest SAScore product in each non-trivial cluster ($1784$ targets left).
    \item Filter out products for which shallow search using Retro*~\cite{chen2020retro} (depth of $6$ nodes, equivalent to $3$ reactions) with the NeuralSym model can find any routes in one minute ($951$ targets left).
\end{itemize}

We then randomly split the resulting hard targets into $151$ targets for validation and $800$ for testing.
Simple random split was justified as due to the clustering any two targets at this stage had fingerprint similarity below $0.87$.

\paragraph{Hyperparameter tuning}
We found that varying the policy temperature $T$ can have a large effect on the behaviour of Retro*, with low temperatures promoting deep greedy exploration of the few most likely steps, while higher temperatures leading to a balanced exploration closer to a breadth-first search.
To ensure a fair comparison, for each model we first ran $10$-minute searches on the $151$ validation targets with $T$ sampled approximately uniformly in log-scale i.e. $T \in \{0.25, 0.35, 0.5, 0.71, 1.0, 1.41, 2.0, 2.83, 4.0\}$.
We then computed solve rate at the $30$, $60$, $120$, $300$ and $600$ second mark, and for each model selected the value of $T$ yielding largest area under the solve rate curve.
We used this setting to produce the final results on $800$ test targets shown in Figure~\ref{fig:search_solve_rate}.

\subsection*{Assessment by domain experts}

The study participants were $9$ PhD-level organic chemists ($4$ at Microsoft and $5$ from major pharmaceutical companies), with a track record of publications and several years of hands-on experience in synthetic organic chemistry.
We first collected outputs on the Pistachio test set from five sources: dataset ground truth, our models (\locmodel, \smilesmodel~and \combinedmodel), and a dummy baseline which applies only rare reaction templates (omitting the most common $4000$) without any ranking.
This allows to compare between our models to ground truth, as well as ground the results in a null baseline which, despite respecting basic syntactic rules due to the use of templates, achieves close to zero recall and leads to mostly nonsensical suggestions which an attentive chemist should be able to spot.
For every pair of sources we sample several test products, and for each consider the top 15 model predictions, only comparing between predictions at the same rank. Cases when the two predictions from the two sources are the same are discarded.

\begin{figure*}[t]
    \centering
    \includegraphics[width=0.8\textwidth]{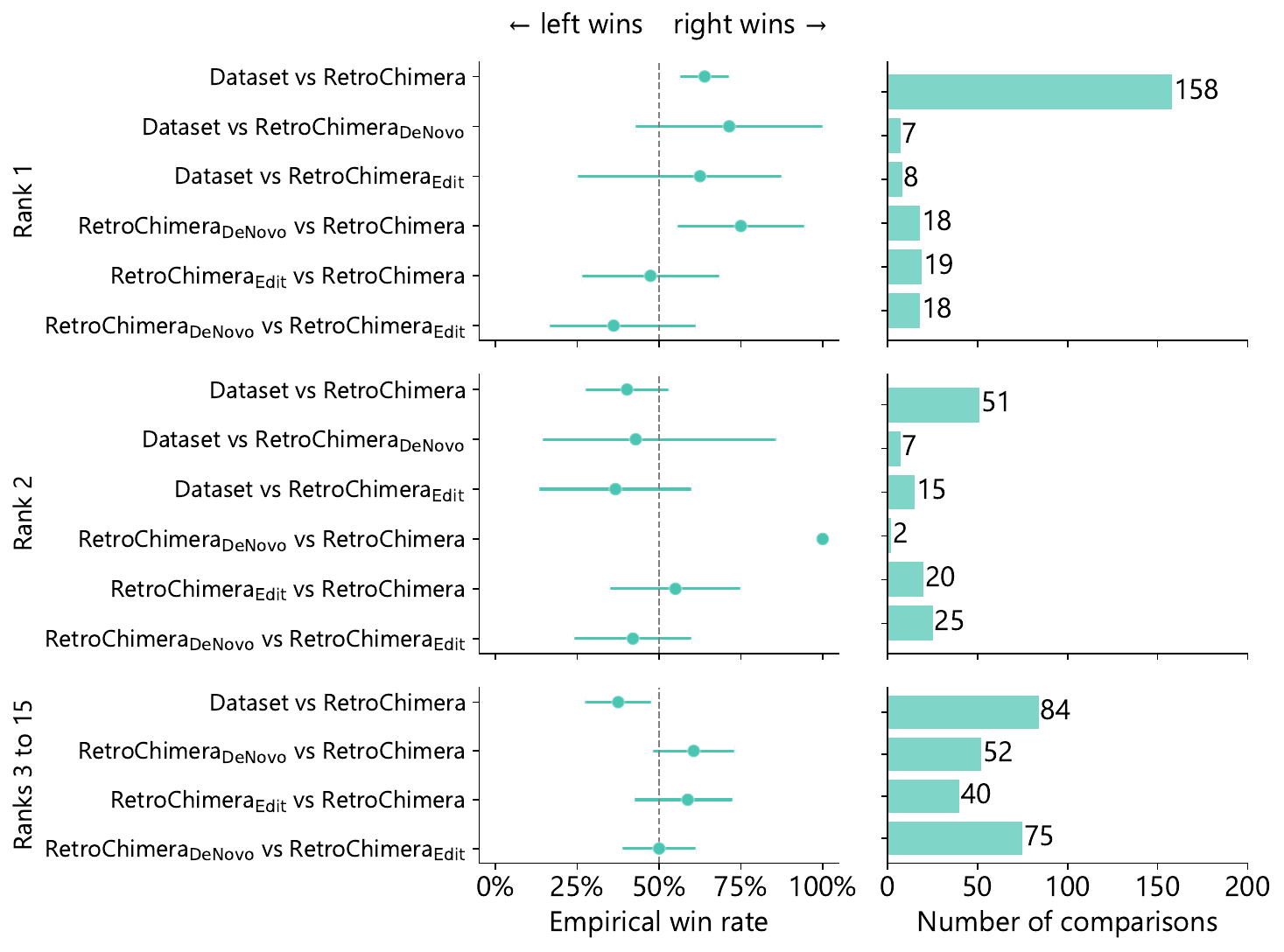}
    \caption{Raw win rate (left) and number of scored pairs (right) describing the data used for the analysis shown in Figure~\ref{fig:preference}. Whiskers next to each win rate correspond to 95\% confidence interval computed using an exact binomial test, which take into account only results for a particular rank bucket and pairs of sources. Note that in some cases the result is not significant due to low number of pairs, but a joint analysis (Figure~\ref{fig:preference}) leads to improved statistical significance.}
    \label{fig:preference_raw}
\end{figure*}

Given a dataset of pairs of predictions, we ask the chemists to judge which one they prefer.
They were given no indication as to the source of each reaction, and order within the pairs was randomized to remove bias.
Coverage of different ranks and pairs of sources was not uniform, chosen to focus on important cases such as \combinedmodel~vs ground-truth (Extended Data Figure~\ref{fig:preference_raw}).
We used comparisons against the dummy baseline to confirm that raters are attentive, but not in the following analysis.

To summarize the preference data, we group by rank (rank 1, rank 2, and ranks 3 through 15). and use the Bradley-Terry model to estimate scores $s_i$ that fit pairwise win rates:

\begin{equation}\label{eq:bradley_terry}
P(\text{source } i \text{ wins with source } j) \approx \frac{e^{s_i}}{e^{s_i} + e^{s_j}}
\end{equation}

We count a tie (i.e. chemist rating both predictions as good, or both as bad) as half a win for both sources.
As we only score pairs where the predictions from the two sources are distinct, observed win rates focus on the cases where the sources diverge, which can be a minority; we found that computing Bradley-Terry scores directly can be sensitive to the distribution of source pairs.
To address this, given the number of scored pairs and the empirical agreement frequency for a given rank bucket and pair of sources, we determine the expected number of agreement cases, and include these as additional ties before computing the Bradley-Terry scores.
Finally, for each model we compute the probability of winning with the dataset (Equation~\ref{eq:bradley_terry}), and use the aforementioned agreement frequency to convert this to win rate conditioned on the predictions being different (Figure~\ref{fig:preference}).

\subsection*{Compute requirements}

Both of our backward models (\locmodel~and \smilesmodel), as well as the forward model based on \smilesmodel, took up to a week to train on a single node with 4-8 A100 GPUs.
The feasibility model was trained in a few days using a single A100 GPU.
Ensembling was done on CPU based on saved model outputs for the underlying models; this allows for learning $\theta$ and evaluating the ensemble without the need to run inference of the original models.

Each search experiment was parallelized over 4-8 GPUs, with each GPU responsible for a subset of targets; we used V100 GPUs for the SimpRetro benchmark and A100 GPUs for our new benchmark based on Pistachio.

\clearpage

\renewcommand{\tablename}{Extended Data Table}
\renewcommand{\figurename}{Extended Data Figure}

\begin{figure*}[t]
    \centering
    \sidesubfloat[]{\hspace{-0.2cm}\includegraphics[height=\plotheight]{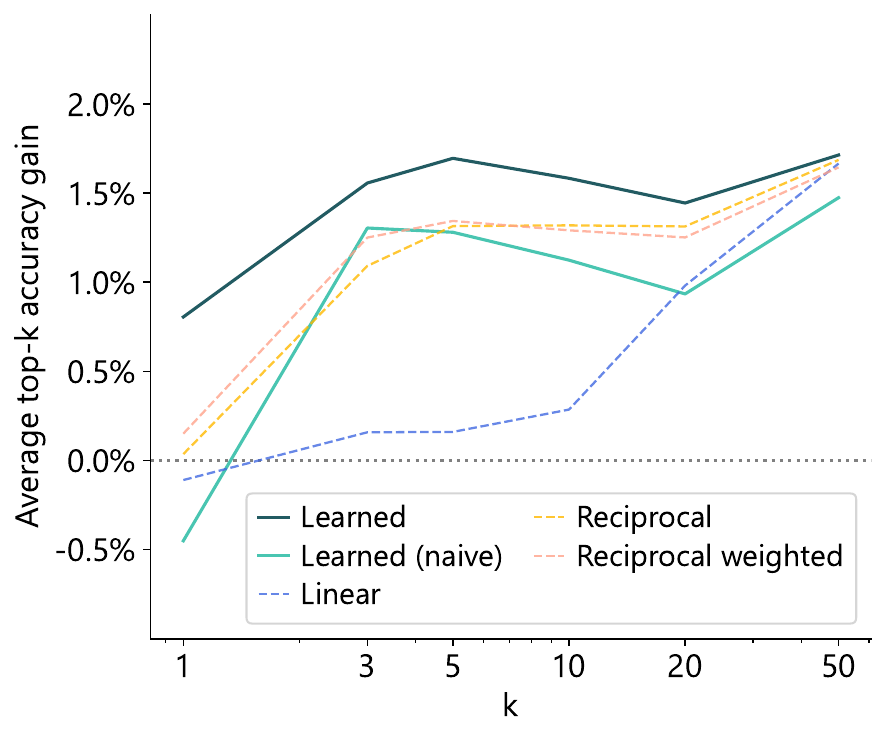}}
    \hfil
    \hspace{0.1cm}
    \sidesubfloat[]{\hspace{-0.2cm}\includegraphics[height=\plotheight]{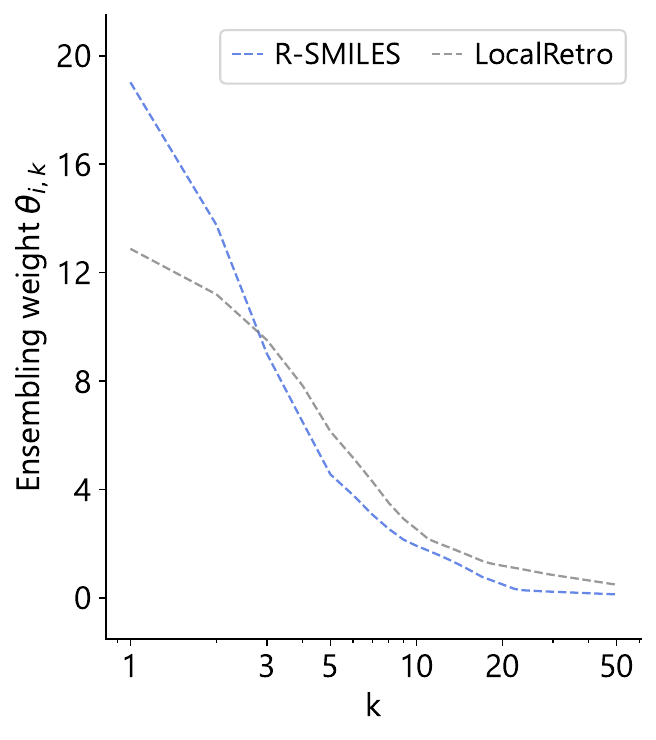}}
    \caption{\textbf{a}, Ablation study for ensembling weight optimization on USPTO-50K. We consider the same models as in Extended Data Figure~\ref{fig:ensemble} together with \locmodel~and \smilesmodel, yielding a total of $11$ models. For every $k$, we show average accuracy gain (over all $55$ model pairs) compared to a baseline formed by taking maximum accuracy among the models in the pair. Our proposed constrained optimization method performs better than a naive approach (no monotonicity or convexity constraints, $w_{reg} = 0$), and several hand-designed weighting schemes: linear ($\theta_{i,k} = k_{max} + 1 - k$), reciprocal ($\theta_{i,k} = \frac{1}{k}$), and weighted reciprocal ($\theta_{i,k} = \frac{c_i}{k}$ where $c_i$ is set to 2 for the model with higher top-1 accuracy and 1 for the weaker model).
    \textbf{b}, Learned weights for combining R-SMILES and LocalRetro on USPTO-50K. We see that the curves cross: R-SMILES is assigned higher weight than LocalRetro for $k \leq 2$ but lower for larger $k$. This highlights that it is not enough to learn the relative model strengths without dependence on rank. We find a similar trend whenever ensembling a de-novo model with an edit-based one.}
    \label{fig:ensemble_analysis}
\end{figure*}

\clearpage

\clearpage

\begin{table*}[h]
\caption{Results on the USPTO-50K dataset with reaction class unknown. Models are grouped by type denoted via the icon on the left: edit-based (\CIRCLE), de-novo (\Circle), and ensemble (\astrosun). Within groups models are sorted by top-1 accuracy. Best result for each top-k accuracy is shown in bold; results that are best within a model type but not best overall are underlined. Results marked with $^*$ utilize techniques proposed in this paper, those marked with $^\dagger$ are taken from prior work, and others were computed using \syntheseus~\cite{maziarz2023re}.}
\label{tab:uspto-50k}
\centering
\begin{tabular}{clcccccc}
\toprule
& Model & top-1 & top-3 & top-5 & top-10 & top-20 & top-50 \\
\midrule
\multirow{8}{0.3cm}{\CIRCLE} & NeuralSym & 45.6\% & 68.1\% & 75.5\% & 82.5\% & 87.9\% & 92.7\% \\
 & MEGAN & 48.7\% & 72.3\% & 79.5\% & 86.7\% & 90.9\% & 93.5\% \\
 & LocalRetro & 51.5\% & 76.5\% & 84.3\% & 91.0\% & \underline{95.0\%} & \underline{96.7\%} \\
 & GLN & 52.4\% & 74.6\% & 81.2\% & 88.0\% & 91.8\% & 93.1\% \\
 & \combinedmodel$_{\text{Edit}}$ (\locmodel)$^*$ & 53.3\% & 74.1\% & 80.7\% & 87.1\% & 91.6\% & 93.8\% \\
 & Graph2Edits & 54.6\% & 76.6\% & 82.8\% & 88.7\% & 91.1\% & 91.7\% \\
 & RetroKNN & 55.3\% & 77.9\% & \underline{85.0\%} & \underline{91.5\%} & 94.8\% & 96.6\% \\
 & RetroExplainer$^\dagger$ & \underline{57.7\%} & \underline{79.2\%} & 84.8\% & 91.4\% & - & - \\
\midrule
\multirow{4}{0.3cm}{\Circle} & Chemformer & 55.0\% & 70.9\% & 73.7\% & 75.4\% & 75.9\% & 76.0\% \\
 & R-SMILES & 56.0\% & 79.1\% & 86.1\% & 91.0\% & 93.3\% & 94.2\% \\
 & EditRetro$^\dagger$ & \textbf{60.8\%} & \underline{80.6\%} & 86.0\% & 90.3\% & - & - \\
 & \combinedmodel$_{\text{DeNovo}}$ (\smilesmodel)$^*$ & 56.9\% & 79.9\% & \underline{86.9\%} & \underline{92.3\%} & \underline{95.5\%} & \underline{96.4\%} \\
\midrule
\multirow{3}{0.3cm}{\astrosun} & \combinedmodel$^*$ & 56.7\% & 80.7\% & 87.6\% & 93.2\% & 96.3\% & 97.9\% \\
 & Ensemble of baselines$^*$ & 59.3\% & 82.3\% & 89.0\% & 94.1\% & 97.0\% & \textbf{98.6\%} \\
 & \combinedmodel++$^*$ & \underline{59.6\%} & \textbf{82.8\%} & \textbf{89.2\%} & \textbf{94.2\%} & \textbf{97.2\%} & \textbf{98.6\%} \\
\bottomrule
\end{tabular}
\end{table*}

\begin{table*}[t]
\caption{Results on the USPTO-FULL dataset, following the same format as Extended Data Table~\ref{tab:uspto-50k} above. Note that RetroWISE was pretrained on additional synthetic data; our understanding of the original work of Zhang et al.\cite{zhang2024retrosynthesis} is that this data was created based on USPTO, thus it may be fair to compare RetroWISE with other models trained on USPTO-FULL. We were not able to confirm this due to the exact code and data not being open-source.}
\label{tab:uspto-full}
\centering
\begin{tabular}{clcccccc}
\toprule
& Model & top-1 & top-3 & top-5 & top-10 & top-20 & top-50 \\
\midrule
\multirow{3}{0.3cm}{\CIRCLE} & LocalRetro$^\dagger$ & 39.1\% & 53.3\% & 58.4\% & 63.7\% & 67.5\% & 70.7\% \\
 & NeuralSym & 44.1\% & 61.4\% & 66.6\% & \underline{71.5\%} & 74.6\% & 77.1\% \\
 & \combinedmodel$_{\text{Edit}}$ (\locmodel)$^*$ & \underline{46.2\%} & \underline{62.0\%} & \underline{66.7\%} & 71.2\% & \underline{74.7\%} & \underline{77.7\%} \\
\midrule
\multirow{4}{0.3cm}{\Circle} & R-SMILES$^\dagger$ & 48.9\% & 66.6\% & 72.0\% & 76.4\% & 80.4\% & 83.1\% \\
 & EditRetro$^\dagger$ & 52.2\% & 67.1\% & 71.6\% & 74.2\% & - & - \\
 & \combinedmodel$_{\text{DeNovo}}$ (\smilesmodel)$^*$ & 51.1\% & 68.1\% & 73.3\% & \underline{78.2\%} & \underline{81.6\%} & \underline{84.8\%} \\
 & RetroWISE$^\dagger$ & \textbf{52.3\%} & \underline{68.7\%} & \underline{73.5\%} & 77.9\% & 80.9\% & 83.6\% \\
\midrule
\multirow{1}{0.3cm}{\astrosun} & \combinedmodel$^*$ & 51.4\% & \textbf{69.5\%} & \textbf{74.6\%} & \textbf{79.5\%} & \textbf{82.8\%} & \textbf{85.6\%} \\
\bottomrule
\end{tabular}
\end{table*}

\clearpage

\begin{figure*}[t]
    \sidesubfloat[]{\includegraphics[height=\plotheight]{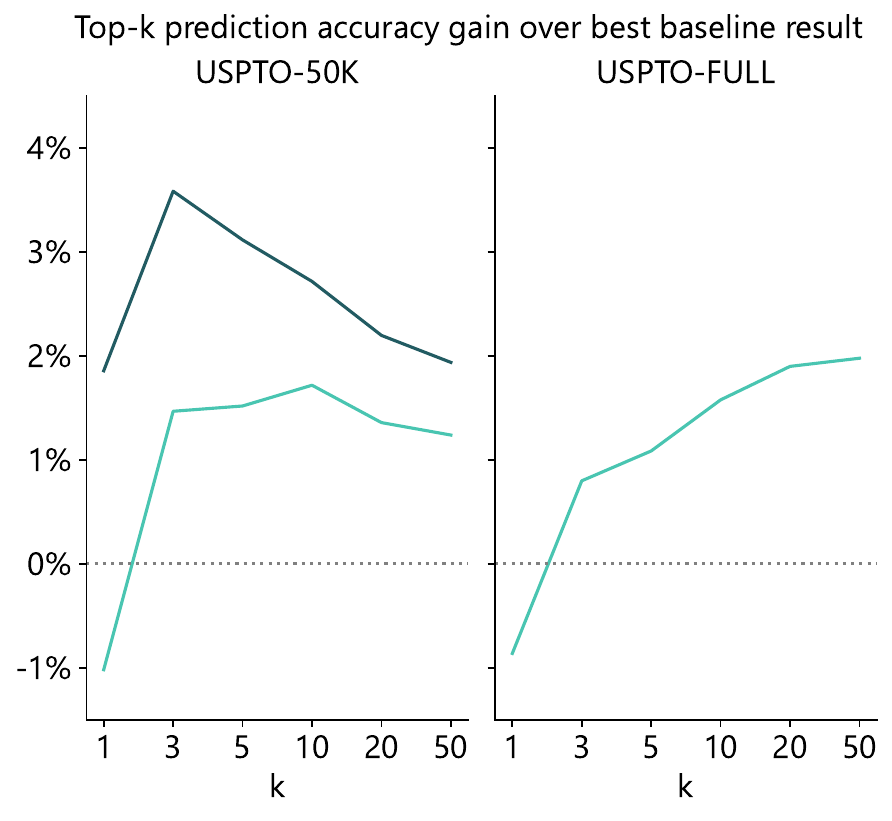}
    \label{fig:accuracy_uspto}}
    \sidesubfloat[]{\includegraphics[height=\plotheight]{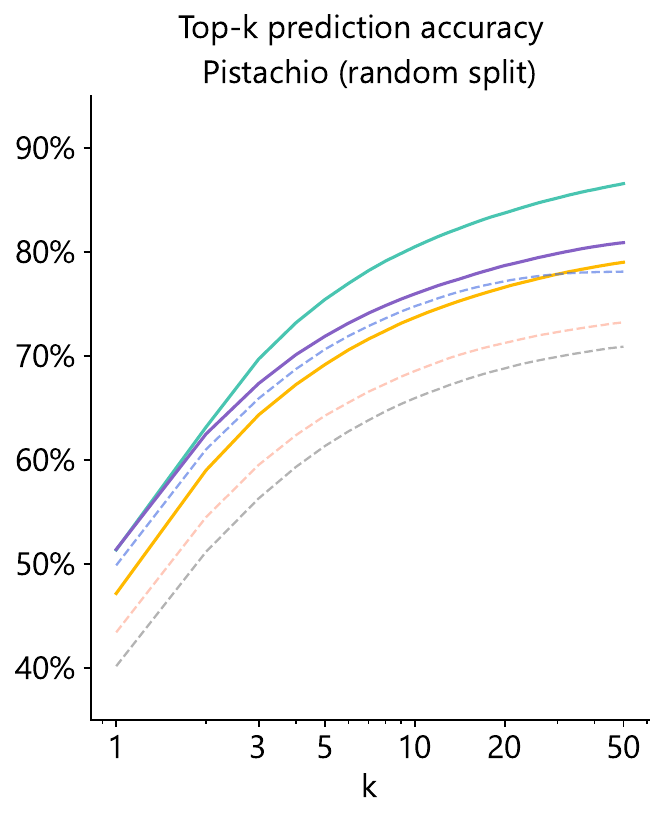}
    }\\
    \vspace{0.1cm}
    \includegraphics[height=0.548cm]{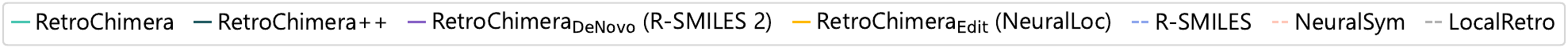}
    \caption{\textbf{a}, Accuracy on USPTO-50K (left) and USPTO-FULL (right), shown as improvement over the best baseline result (selected for each $k$ separately). \combinedmodel++ is an ensemble of both our models and baselines ($m = 10$). \textbf{b}, Accuracy on the random split test set of Pistachio proposed by Maziarz et al\cite{maziarz2023re}. Some performance differences are accentuated compared to our time-split test set, but the model ranking is largely preserved.}\label{fig:accuracy_pistachio_random}
\end{figure*}

\clearpage

\clearpage

\begin{table*}[h]
\caption{Architectural, training and inference hyperparameters of the \locmodel~model across the datasets investigated in this work.}
\label{tab:locmodel-arch-hparams}
\centering
\begin{tabular}{llccc}
\toprule
& Parameter & USPTO-50K & USPTO-FULL & Pistachio \\
\midrule
& $d_{\text{clf}}$ & 256 & 256 & 256 \\
& $d_{\text{free}}$ & 0 & 32 & 32 \\
& Number of templates & $9735$ & $228\,127$ & $146\,256$ \\
\midrule
\multirow{8}{1.3cm}{$\text{GNN}^{\text{in}}$}
& Layer type & GPS + PNA & PNA & PNA \\
& Number of layers & 3 & 5 & 5 \\
& Hidden dim & 64 & 768 & 1024 \\
& Output dim (node-level) & 256 & 128 & 128 \\
& Output dim (graph-level) & 512 & 1024 & 1024 \\
& Aggregation heads & 8 & 8 & 8 \\
& Dropout (inter-layer) & $0.1$ & $0.0$ & $0.05$ \\
& Dropout (post aggregation) & $0.4$ & $0.4$ & $0.4$ \\
\midrule
\multirow{8}{1.3cm}{$\text{GNN}^{\text{tpl}}$} 
& Layer type & GPS + PNA & PNA & PNA \\
& Number of layers & 4 & 5 & 5 \\
& Hidden dim & 64 & 192 & 192 \\
& Output dim (node-level) & 256 & 128 & 128 \\
& Output dim (graph-level) & 512 & 512 & 512 \\
& Aggregation heads & 8 & 8 & 8 \\
& Dropout (inter-layer) & $0.1$ & $0.0$ & $0.0$ \\
& Dropout (post aggregation) & $0.4$ & $0.4$ & $0.4$ \\
\midrule
& Batch size & 128 & 256 & 512 \\
& Number of epochs & 600 & 130 & 85 \\
& Initial learning rate & $10^{-3}$ & $10^{-3}$ & $10^{-3}$ \\
& Loss type & softmax & sigmoid & sigmoid \\
& $r^{\text{clf}}$ & - & 30 & 18 \\
& $r^{\text{loc}}$ & 1 & 4 & 4 \\
& $r^{\text{app}}$ & 100 & 10 & 10 \\
\midrule
& Total parameter count & 1.9M & 103M & 165M \\
\bottomrule
\end{tabular}
\end{table*}

\begin{table*}[h]
\caption{Architectural, training, and inference hyperparameters of the \smilesmodel~model across the datasets investigated in this work.}
\label{tab:smimodel-arch-hparams}
\centering
\begin{tabular}{llccc}
\toprule
& Parameter & USPTO-50K & USPTO-FULL & Pistachio \\
\midrule
& Vocab size & 72 & 235 & 346 \\
& Number of layers & 6 & 6 & 8 \\
& Hidden dim & 256 & 512 & 512 \\
& Feedforward dim & 512 & 2048 & 2048 \\
& Number of heads & 8 & 8 & 8 \\
& Number of KV heads & 8 & 2 & 2 \\
\midrule
& Batch size & 128 & 128 & 512 \\
& Number of epochs & 30 & 60 & 30 \\
& Learning rate scheduler & Noam & Noam & Noam \\
& Learning rate & 1.0 & 1.0 & 1.0 \\
& Warmup steps & 8000 & 8000 & 8000 \\
& Dropout & 0.3 & 0.1 & 0.1 \\
\midrule
& Number of augmentations & 20 & 5 & 10 \\
& Beam size & 10 & 50 & 20 \\
\midrule 
& Total parameter count & 17.4M & 44.5M & 66.7M \\
\bottomrule
\end{tabular}
\end{table*}

\end{document}